\documentclass[sigconf,nonacm]{acmart}
\AtBeginDocument{%
  \providecommand\BibTeX{{%
    \normalfont B\kern-0.5em{\scshape i\kern-0.25em b}\kern-0.8em\TeX}}}

\usepackage{multirow}
\usepackage{subcaption}
\usepackage{amsfonts,mathtools}

\usepackage{fixmath}

\begin{document}
\title{Time Series Language Model for Descriptive Caption Generation
}

\author{Mohamed	Trabelsi}
\email{mohamed.trabelsi@nokia-bell-labs.com}
\affiliation{%
 \institution{Nokia Bell Labs}
  \city{Murray Hill}
  \state{NJ}
  \country{USA}
}

\author{Aidan Boyd}
\email{aidan.boyd@nokia-bell-labs.com}
\affiliation{%
 \institution{Nokia Bell Labs}
  \city{Murray Hill}
  \state{NJ}
  \country{USA}
}

\author{Jin Cao}
\email{jin.cao@nokia-bell-labs.com}
\affiliation{%
 \institution{Nokia Bell Labs}
  \city{Murray Hill}
  \state{NJ}
  \country{USA}
}

\author{Huseyin Uzunalioglu}
\email{huseyin.uzunalioglu@nokia-bell-labs.com}
\affiliation{%
  \institution{Nokia Bell Labs}
  \city{Westford}
  \state{MA}
  \country{USA}
}

\begin{abstract}

The automatic generation of representative natural language descriptions for observable patterns in time series data enhances interpretability, simplifies analysis and increases cross-domain utility of temporal data. While pre-trained foundation models have made considerable progress in natural language processing (NLP) and computer vision (CV), their application to time series analysis has been hindered by data scarcity. 
Although several 
large language model (LLM)-based methods have been proposed for time series forecasting, time series captioning is under-explored in the context of LLMs. In this paper, we introduce TSLM, a novel time series language model designed specifically for time series captioning. TSLM operates as an encoder-decoder model, leveraging both text prompts and time series data representations to capture subtle temporal patterns across multiple phases and generate precise textual descriptions of time series inputs. 
TSLM addresses the data scarcity problem in time series captioning by first leveraging an in-context prompting synthetic data generation, and second denoising the generated data via a novel cross-modal dense retrieval scoring applied to time series-caption pairs. Experimental findings on various time series captioning datasets demonstrate that TSLM outperforms existing state-of-the-art approaches from multiple data modalities by a significant margin.
  
\end{abstract}




\maketitle

\section{Introduction}
Time series analysis \cite{Dama2021TimeSA,HornMBRB20} is a fundamental problem with significant applications in various real-world cases \cite{wen_kdd_22}. It plays a crucial role in multiple tasks such as forecasting \cite{jin2024timellm,chang2024llm4ts,zhou2023onefitsall} in several domains including energy consumption \cite{ZhouZPZLXZ21}, weather \cite{abs-2401-00230}, disease propagation \cite{MatsubaraSPF14}, traffic \cite{abs-2401-00230}, finance \cite{YanO18}, anomaly detection \cite{abs-2211-05244,abs-2402-02032}, etc. The automatic generation of natural language descriptions for salient patterns in time series data \cite{jhamtani2021truth,MurakamiWMGYTM17,SowdaboinaCS14,MahinpeiKT22} is an important task that significantly enhances the overall process of time series analysis. Translating complex numerical data into understandable narratives improves the interpretability of data, making it accessible to a broader audience including those without specialized knowledge in data science or statistics.
Furthermore, it facilitates more effective communication of insights derived from time series analysis, allowing individuals and organizations to quickly grasp key trends, anomalies, and other significant patterns. Automating the description of time series patterns streamlines the analytical process and reduces the time and effort required to extract meaningful insights from large datasets. For instance, in finance, automatically generated descriptions can help analysts quickly identify market trends and inform investment strategies. The value of the time series captioning comes from captioning at scale. Given a large number of time series instances, time series captioning models can automatically caption these time series and then use these captions for downstream tasks such as summarization, statistics generation, and trend analysis. Attaining human annotations for large numbers of time series is infeasible, whereas time series captioning models enable this in an effective and efficient manner. In healthcare, time series captioning models can assist in monitoring patient vitals and predicting potential health issues. These automatically generated interpretations from time series allow the medical staff to make informed decisions without the need for manual inspections of the time series graphs. In manufacturing, time series captioning models can be used to detect anomalies in production processes and prevent equipment failures.

Despite the extensive research and development in time series analysis methodologies, the domain of time series captioning remains significantly under-explored, particularly in the context of large language models (LLMs). Time series forecasting has seen numerous advancements, with several techniques \cite{jin2024timellm,chang2024llm4ts,zhou2023onefitsall,liang_time_series_analysis_survey,timegpt,abs-2310-10688,abs-2402-02592} being employed to predict future data points based on historical patterns. However, the task of generating natural language descriptions that summarize and elucidate the key patterns and insights from time series data has not received comparable attention. LLMs exhibit robust pattern recognition skills when processing complex sequences of tokens, therefore the advanced natural language understanding and generation aspects of LLMs can be leveraged to create detailed, coherent, and contextually accurate descriptions of time series data. By incorporating contextual and semantic nuances into the generated captions, LLMs provide deeper insights and more meaningful interpretations compared to traditional methods \cite{jhamtani2021truth,MurakamiWMGYTM17,SowdaboinaCS14,MahinpeiKT22}. In addition, recent work \cite{llava,flamingo,multimodal_gpt,audiogpt} has shown impressive results for multi-modal generative AI which integrates and processes multiple types of data such as text, images, audio, and video. 
By combining different data modalities, multi-modal generative AI achieves a more comprehensive understanding of context, leading to more accurate content generation. In addition, multi-modal generative AI extracts richer representations of data by capturing complementary information from different modalities of data.

Inspired by the recent progress of LLMs and multi-modal generative AI, we propose a new multi-modal model, called \textit{\textbf{T}ime \textbf{S}eries \textbf{L}anguage \textbf{M}odel} (TSLM), that consumes a joint representation of text and embedding for a time series, and outputs a textual description of a time series. A time series can be seen as both (1) a sequence of numerical values; and (2) a sequence of tokens where each token is the string representation of the corresponding numerical value. We propose to form a textual representation of the time series by explicitly injecting positional information into the text-based sequence with phase tagging leading to a coarse-grained sequence representation. To capture fine-grained information about subtle variations and intricate patterns in a time series, we propose to embed the sequence of numerical values of a time series using a time series 1D CNN-based encoder. This encoder is trained in an autoencoder setup to learn efficient representation of the time series data without the need for labels in an unsupervised learning setting. The coarse- and fine-grained representations of a time series are combined into a joint representation that is used as input to the multi-modal encoder. To achieve better modality alignment between the text and time series within the multi-modal encoder, we incorporate a reprogramming layer \cite{jin2024timellm} to reprogram the time series embeddings into the textual representation space. The aligned time series embeddings are concatenated with the textual embeddings to form the textual- and time series-aware embeddings that are forwarded to the transformer blocks of our multi-modal encoder, which outputs the final self-attention embeddings that encode both textual- and time series-based embeddings. 

Public time series captioning datasets are scarce which makes training LLMs ineffective and risks overfitting. Therefore, we propose an in-context prompting data generation to augment the training dataset through few-shot learning from high-quality demonstrations. Our data generation process is based on open-source LLMs so that we can generate large quantities of data without incurring financial costs. This synthetically generated data may contain plausible but incorrect factual information as a result of hallucinations. Therefore, we propose a novel denoising method of the generated data via cross-modal dense retrieval scoring. We train a cross-modal dense retrieval scoring model using the original clean data, where the objective is to enhance the understanding of the relationship between the joint representation of the time series and the textual representation of the caption. Then, the trained cross-modal dense retrieval scoring is used to filter noisy time series-caption pairs from the generated data by assessing the similarity between modalities, and pairs with low computed similarity from the generated data can be denoted as noisy. The original and denoised generated data are used to train our new encoder-decoder TSLM with the next token prediction task. In the generation phase, TSLM receives an unseen time series and generates multiple captions that are summarized by an existing LLM to obtain a single descriptive caption. Therefore, our TSLM acts as a bridge between the time series data and the LLM-based summarizer. 

In summary, we make the following contributions: 

\begin{itemize}
    \item We propose a new multi-modal model, called Time Series Language Model (TSLM), that consumes both textual- and embedding-based representations of a time series to generate accurate textual descriptions of a time series.
    \item We propose an in-context prompting synthetic data generation for time series captioning by leveraging open-source LLMs in order to augment the training dataset with large quantities of samples without incurring financial costs.
    \item  We propose a new denoising method of the generated data via cross-modal dense retrieval scoring which is used to filter time series-caption pairs from the generated data by capturing the similarity between the joint representation of the time series and the textual representation of the caption.
    \item We evaluate over two datasets, and demonstrate that our new method
outperforms the state-of-the-art baselines from multiple modalities.

\end{itemize}
\section{Related work}

\subsection{Time Series Captioning}
The task of time series captioning \cite{jhamtani2021truth,MurakamiWMGYTM17,SowdaboinaCS14,MahinpeiKT22} refers to generating natural language descriptions that summarize and elucidate the key patterns and insights from time series data. Murakami et al. \cite{MurakamiWMGYTM17} proposed to generate market comments from stock prices with an encoder-decoder model, where several encoding methods, including Multi-Layer Perceptron (MLP), Convolutional Neural Network (CNN), or Recurrent Neural Network (RNN) with Long Short-Term Memory (LSTM) cells
, are tested. Sowdaboina et al. \cite{SowdaboinaCS14} addressed the task of describing wind speed and direction by selecting important points in a time series that can help in caption generation. The authors use simple statistical measures and characteristic features that are derived from the wind direction and speed 
to build a simple MLP-based classifier. Jhamtani et al. \cite{jhamtani2021truth} proposed a neural truth-conditional model for time series captioning. This model learns to identify patterns that are relevant to a time series by leveraging three simple types of modules: \textit{pattern}, \textit{locate}, and \textit{combine} defined as neural modular networks \cite{AndreasRDK16}. These modules are composed together to form handcrafted programs so that the caption generation is conditioned on only the logical program to generate an output text that describes this pattern via a decoder.

Time series data in the form of charts has been used in the task of figure question answering \cite{KafleSPCK20}. Time series data are considered as line charts for captioning in the method proposed by Mahinpei et al. \cite{MahinpeiKT22}. This model adapts the PReFIL \cite{KafleSPCK20} model for line charts captioning. Similar to image captioning, this model is composed of a DenseNet \cite{densenet} to process the figure image, two fusion blocks for processing high- and low-level feature maps from the DenseNet, an LSTM to process the figure’s caption one token at a time, and a neural network classifier that predicts the next token.

\subsection{Multi-modal Models}

Deep contextualized language models, such as BERT \cite{Devlin2019BERTPO} and RoBERTa \cite{Liu2019RoBERTaAR}, have been proposed to solve multiple tasks in information retrieval (IR) \cite{sakata2019,dai2019,strubert,Chen2020TableSU,survey_doc_retrieval,khattab_sigir,nogueira_passage,nogueira_multi} and NLP \cite{WangHCS20,dame,selab,selab_arxiv,WaddenWLH19,ZhangWZ19,absformer}. Recently, researchers have focused on the Generative Pre-trained Transformer (GPT) models to advance LLM capabilities in multiple tasks \cite{SunL0WGZ023,loggpt,WangYW24,zhang2023,MinRSVNSAHR24,abs-2209-12356}. In this regard, 
multi-modal models have been developed in the context of LLMs to improve the understanding and content generation across various modalities such as text, images, audio, and video. CLIP \cite{clip} addressed the alignment of visual and textual representations by learning from image-description pairs in order to bridge the gap between language and vision. The fusion of visual and textual data has led to the creation of vision-language models (VLMs), which excel at a range of tasks including image captioning, visual question answering (VQA), and generating images from textual descriptions. 
For example, the DALL-E \cite{dalle} model generates high-quality images from textual inputs by extending the capabilities of LLMs. Flamingo \cite{flamingo} is another visual language model that can perform various multi-modal tasks such as captioning, visual dialogue and VQA 
using only a few input/output samples. Flamingo can efficiently use as inputs arbitrarily interleaved text and visual data input and generate text as output by inserting gated cross-attention dense blocks between the original LLM layers, that are trained from scratch.
MultiModal-GPT \cite{multimodal_gpt} adapts the Flamingo model to build a vision and language model for dialogue with humans where LLaMA is used as the language decoder. Multi-modal models have been studied in the context of instruction-following agents through visual instruction tuning, where LLaVA \cite{llava} has shown impressive results in instruction-following and visual reasoning capabilities.

Other combinations of modalities include audio and textual modalities, where AudioGPT \cite{audiogpt} has been proposed to understand and generate audio content from textual input for applications related to speech recognition, audio generation, and sound extraction and detection.

Cross-modal retrieval has also benefited from the advancements in multi-modal models. For example, Florence \cite{florence} and ALIGN \cite{align} have shown the capability to retrieve images based on textual queries and texts based on visual queries.

\section{Problem Statement}

The time series captioning task consists of automatically generating a natural language description $c$ of the time series input $T$, where the goal is to provide meaningful descriptions that capture the essential patterns, trends, or events present in the time series data. Time series captioning involves processing the temporal sequence of data points and producing coherent and informative textual output. 

When training time series captioning models, multiple time series-caption pairs $D = \{(T_1,c_1),(T_2,c_2),\ldots,(T_{|D|},c_{|D|})\}$ are given, where $|D|$ is the total number of pairs.
Given the scarcity of time series-caption pairs, TSLM involves a data generation that consists of generating synthetic data from existing LLMs to augment the training dataset and increase the diversity and size of the dataset, which in turn improves the robustness and 
generalization ability of TSLM. While it is possible to automatically generate a wide range of
synthetic data, the process inevitably introduces noisy time series-caption pairs. Therefore, TSLM involves a new denoising method that is applied to the synthetically generated data, before training TSLM with the resulting denoised data. 

The trained TSLM model is used to generate multiple captions of the time series input, then these captions are summarized using pre-trained LLMs to generate a descriptive caption.

\section{TSLM: Multi-modal Encoder}
In this section, we introduce the multi-modal encoder of TSLM that consumes a joint representation of text and embedding for a time series, and outputs textual- and time series-aware embeddings.
\subsection{Time Series Representations}
A univariate time series $T=t_1t_2\ldots t_l$ is composed of $l$ numerical values corresponding to $l$ different timestamps. We propose a joint representation for a time series that is composed of both textual and embedding representations to cover multiple modalities.
\subsubsection{Textual Representations}\label{TR}
The time series $T$ can be seen as a sequence of tokens where each token is the string representation of the corresponding numerical value. To explicitly inject positional information into the text-based sequence of $T$, we add phase tags to the sequence to distinguish three phases which are: starting, middle, and end. We denoted the tagged time series by $<time\_series>$ and is given by:
\begin{equation} 
\label{text-rep}
\begin{split}
<time\_series> \textrm{ } & = \textrm{ }<start> \textrm{ }T_{1:\frac{l}{3}} \textrm{ }</start> \\
 & + \textrm{ }<middle> \textrm{ }T_{\frac{l}{3}+1:\frac{2l}{3}} \textrm{ }</middle> \\
 & + \textrm{ }<end> \textrm{ }T_{\frac{2l}{3}+1:l} \textrm{ }</end>
\end{split}
\end{equation}
where $T_{i:j}, i<j$ denotes the subsequence $t_it_{i+1}\ldots t_j$ of $T$, $+$ denotes the string concatenation operation, $\{<start>,</start>\}$, $\{<middle>,</middle>\}$, and $\{<end>,</end>\}$ denote the tags of the start, middle, and end phases, respectively. The textual representation of $T$ captures a position-aware coarse-grained information that is beneficial for the downstream task of the time series captioning. In general, multiple tagging can be used to segment the time series sequence. For this work, we only focus on the three phases segmentation. We leave investigating more sophisticated segmentation techniques as a future work.

\subsubsection{Embedding Representations}
The time series $T$ can be seen as a sequence of numerical values. The time series embeddings encode fine-grained information about the behavior of the time series. This allows the model to capture subtle variations and intricate patterns in the data. 
In addition, the time series encoder can reduce the dimensionality of the data, making it easier to handle variable length sequences and process long sequences. Therefore, the time series encoder provides a compressed and informative representation of the time series that is beneficial for the downstream task of the time series captioning. $T$ is encoded into an embedding, denoted by $<time\_series\_embedding>$, using a time series encoder:
\begin{equation} 
\label{embedding}
<time\_series\_embedding> = Time Series Encoder (T) \in \mathbb{R}^{f \times d}
\end{equation}
where $f$ is the size of the compressed time series, and $d$ is the dimension of the embedding. 

\subsubsection{Joint Representations}
We propose to represent a time series $T$ by combining both coarse- and fine-grained information captured by the textual- and embedding-based representations, respectively. This leads to a joint representation, denoted by $JR$, in the context of time series captioning that is given by:
\begin{equation} 
\label{joint}
\begin{split}
JR(T)& = [CLS] \textrm{ }Describe\textrm{ }this\textrm{ }time\textrm{ }series\textrm{ }<time\_series> \\
 & \textrm{ }\textrm{ }\textrm{ }\textrm{ }\textrm{ }\textrm{ }\textrm{ }\textrm{ }\textrm{ }\textrm{ }encoded\textrm{ }by\textrm{ }<time\_series\_embedding>
\end{split}
\end{equation}
where $[CLS]$ is a special token that is added to the beginning of the joined representation. The joint  representation $JR(T)$ of the time series $T$ is used as input to the multi-modal encoder.

\subsection{Multi-Modal Encoder Architecture}
Our model is composed of an LLM-based encoder that contains an embedding layer denoted by $embed\_tokens$ and transformer blocks denoted by $transformer\_blocks$. We embed the text part of the joint representation $JR(T)$, denoted by $JR_t(T)$, using the embedding layer:
\begin{equation} 
\begin{split}
JR_t(T)& = [CLS] \textrm{ }Describe\textrm{ }this\textrm{ }time\textrm{ }series\textrm{ }<time\_series> \\
 & \textrm{ }\textrm{ }\textrm{ }\textrm{ }\textrm{ }\textrm{ }\textrm{ }\textrm{ }\textrm{ }\textrm{ }\textrm{ }\textrm{ }\textrm{ }\textrm{ }\textrm{ }\textrm{ }\textrm{ }\textrm{ }\textrm{ }\textrm{ }\textrm{ }\textrm{ }\textrm{ }encoded\textrm{ }by
\end{split}
\end{equation}
\begin{equation} 
\boldsymbol{E_t}=embed\_tokens(JR_t(T)) \in \mathbb{R}^{n \times d}
\end{equation}

where $n$ represents the number of tokens resulting from tokenizing $JR_t(T)$, 
and $d$ is the dimension of the text embedding which is assumed to be equal to the dimension of the time series embeddings.

To achieve better modality alignment between text and time series, we incorporate a reprogramming layer \cite{jin2024timellm} to reprogram the time series embeddings into the textual representation space. Instead of a patch reprogramming as in \cite{jin2024timellm}, our reprogramming layer operates directly on the time series embeddings and outputs a reprogrammed time series embeddings that are more aligned with the textual embeddings $E_t$. The time series embeddings should be aligned with the embedding matrix of the LLM vocabulary denoted by $\boldsymbol{V}\in \mathbb{R}^{|\boldsymbol{V}| \times d}$, with $|\boldsymbol{V}|$ is the size of the vocabulary. This alignment is learned through text prototypes that reduce the computational overhead of the alignment. A parametric linear layer $P \in \mathbb{R}^{p \times |\boldsymbol{V}|}$ ($p<<|\boldsymbol{V}|$) is introduced to obtain the text prototypes embeddings:
\begin{equation} 
\label{embedding}
\boldsymbol{E_p}=P\boldsymbol{V} \in \mathbb{R}^{p \times d}
\end{equation}
To align the time series embeddings with the text prototypes, we introduce a transformer-based cross attention layer that is composed of $H$ heads. For each head $h \in H$, three parametric matrices are introduced: a query matrix $ Q_h\in \mathbb{R}^{d \times d_h}$, a key matrix $K \in \mathbb{R}^{d \times d_h}$, and a value matrix $V \in \mathbb{R}^{d \times d_h}$, where $d_h=\frac{d}{H}$. For a given time series $T$, the cross-attention between the time series embeddings 
and the text prototypes embeddings 
is given by:
\begin{equation}
\begin{array}{l}
\mathcal{Q}_h=<time\_series\_embedding> Q_h \in \mathbb{R}^{f \times d_h} \\
\mathcal{K}_h=\boldsymbol{E_p} K_h \in  \mathbb{R}^{p \times d_h}  \\
\mathcal{V}_h=\boldsymbol{E_p} V_h \in  \mathbb{R}^{p \times d_h} \\
\mathcal{Z}_h=\operatorname{softmax}\left(\frac{\mathcal{Q}_h \mathcal{K}_h^{T}}{\sqrt{d_h}}\right) \mathcal{V}_h \in \mathbb{R}^{f \times d_h} \\
\mathcal{Z} = \mathcal{Z}_1 \oplus \mathcal{Z}_2 \oplus \ldots \oplus \mathcal{Z}_H \in  \mathbb{R}^{f \times d}
\end{array}
\end{equation}
where $\oplus$ denotes the matrix concatenation operation. So, the key output $\mathcal{K}_h$ and the value output $\mathcal{V}_h$ are computed using the text prototypes embeddings, 
and the query output is computed using the time series embeddings. 
The aligned time series embeddings $\mathcal{Z}$ are concatenated with the text embeddings $\boldsymbol{E_t}$ to form the textual- and time series-aware embeddings denoted by $\boldsymbol{E_t^s}$:
\begin{equation}
\boldsymbol{E_t^s} = \boldsymbol{E_t} \oplus \mathcal{Z} \in  \mathbb{R}^{(n+f) \times d}
\end{equation}
The textual- and time series-aware embeddings $\boldsymbol{E_t^s}$ are forwarded to the $transformer\_blocks$ layer to obtain the final self-attention embeddings, denoted by $\boldsymbol{X}$, which encode both textual- and time series-based embeddings :
\begin{equation}
\boldsymbol{X} = transformer\_blocks (\boldsymbol{E_t^s}) \in  \mathbb{R}^{(n+f) \times d}
\end{equation}
We define two varieties of our proposed encoder based on the parts of $\boldsymbol{X}$ that are used by the following components: (1) if the full embedding matrix $\boldsymbol{X} \in  \mathbb{R}^{(n+f) \times d}$ is used, we denote the encoder by \textbf{\textit{Multi-Modal Encoder (Matrix)}}; (2) if only the embedding of the $[CLS]$ token, $\boldsymbol{X}_{[CLS]} \in  \mathbb{R}^d$, is pooled, we denote the encoder by \textbf{\textit{Multi-Modal Encoder (Vector)}}.

\section{TSLM: Training with Denoised Generated Data}
\begin{figure*}[t!]
\centering
\includegraphics[width=1\textwidth]{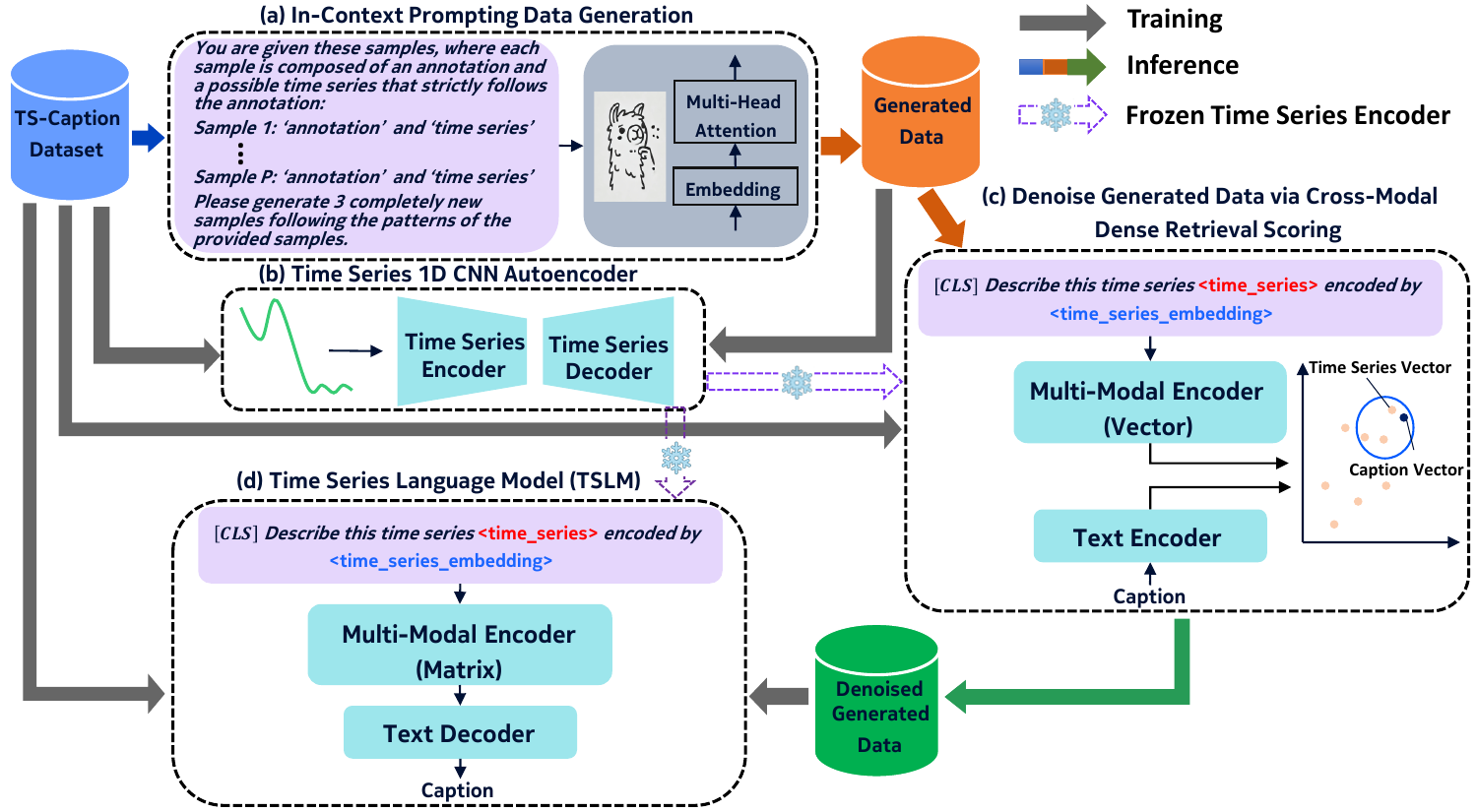}
\caption{The overview of training TSLM, 
which is composed of four key steps: (a) In-context prompting data generation; (b) Time series 1D CNN autoencoder; (c) Denoise generated data via cross-modal dense retrieval scoring; and the final step (d) is the training of TSLM with the denoised generated data.}
\label{training_phase}
\end{figure*}

In this section, we introduce the training steps of TSLM. 
As shown in Figure \ref{training_phase}, the training phase 
is composed of four key 
steps: (1) In-context prompting data generation to generate time series-caption pairs; (2) Times series 1D CNN autoencoder to learn time series embeddings; (3) Denoise generated data via cross-modal dense retrieval scoring to discard noisy generated pairs; and (4) Time series language model where the actual training of TSLM happens.

\subsection{In-Context Prompting Data Generation} \label{data generation}
High-quality and large-scale data is fundamental for training effective LLMs. Public time series captioning datasets are scarce which makes training LLMs ineffective and risks overfitting. Data generation creates additional training samples, which is particularly useful for the robustness and accuracy of time series captioning models. 
Therefore, synthetic data generation addresses the scarcity of such data, enabling more advanced and reliable LLMs by achieving better generalization, reducing overfitting, and improving performance on unseen data.
LLMs have demonstrated impressive zero-shot generalization capabilities across various NLP applications, largely due to extensive pre-training. 
Recent methods \cite{ChiaBPS22,DingQLCLJB23} leverage the implicit 
knowledge embedded in LLMs to generate synthetic data for downstream tasks, as follows:
\begin{equation}
\label{in-context}
[s_i,y_i] = LLM(q_i)
\end{equation}
where $q_i$ is the input query sequence, $s_i$ and $y_i$ are the generated sequence and label by the LLM, respectively. 
In-context prompting 
guides the behavior of LLMs to achieve desired outcomes without modifying the model's weights. This involves few-shot learning by providing a set of high-quality demonstrations, each with both input and desired output, for the target task. By first encountering these examples, the model gains a clearer understanding of human intentions and the criteria for desired answers, typically resulting in better performance compared to zero-shot learning. 
For in-context prompting data generation in time series captioning, $s_i$ and $y_i$ in Equation (\ref{in-context}) refers to the generated time series and caption, respectively. As shown in Figure \ref{training_phase} (a), the query $q_i$ is composed of $P$ demonstrations that are selected from the groundtruth time series captioning dataset denoted by \textit{TS-Caption Dataset}, and the synthetic data generation instruction which instructs the LLM to generate $S$ time series-caption pairs. The time series part of each demonstration follows the tagged text representation as shown in Equation (\ref{text-rep}), therefore the time series part of a generated pair also follows the tagged representation. 

Diversity and quality of the generated data are fundamental criteria in synthetic data generation. To increase the diversity of the generated data, we propose to bootstrap the data generation process by including the already generated data into the groundtruth \textit{TS-Caption Dataset}. Therefore, the final generated data can simulate a wide range of scenarios that may not be present in the original dataset, allowing the LLM to learn from a broader spectrum of examples. The bootstrapping during the synthetic data generation significantly reduces the number of duplicate samples in the generated data. To improve the quality of data generation, we group the demonstrations by computing the string similarity (using fuzzywuzzy\footnote{https://pypi.org/project/fuzzywuzzy/})
among the annotations. This means that the formed groups contain annotations that share similarities, and this helps the LLM to generate better quality data.

We use LLaMA2-13B-Chat\footnote{https://ai.meta.com/llama/} 
as the LLM in Equation (\ref{in-context}) to generate synthetic data. While manual inspection of a few samples suggests that ChatGPT\footnote{https://chat.openai.com/} 
produces higher quality data, we choose the open-source LLaMA2-13B-Chat 
model to generate large quantities of data without incurring financial costs. We leave investigating more open-source models, such as LLaMA 3, 3.1, and 3.2, for synthetic time series captioning data generation as a future work. The in-context prompting data generation step results in a large-scale and diverse data denoted as \textit{Generated Data} in Figure \ref{training_phase}. This data is generated in a cost- and time-effective way by querying the open-source LLaMA model.

\subsection{Time Series 1D CNN Autoencoder}

Convolutional autoencoders \cite{BerahmandDSLX24,LiPL23} learn to encode an input image in a high-dimensional space into a lower-dimensional feature space, which enables dimensionality reduction while preserving important information and discarding noise. These convolutional layers share weights across spatial locations which leads to a more efficient training. In addition, the high-quality spatial features that are learned by the convolutional autoencoders can be incorporated into a multi-modal model. We leverage the power of convolutional autoencoders to learn the time series embeddings as shown in Figure \ref{training_phase} (b). For the scope of this paper, we only consider univariate time series. We introduce 1D CNN layers to capture local patterns and temporal dependencies within a univariate time series. By leveraging 1D CNN layers, we can handle variable length time series sequences. In addition, the 1D CNNs are able to recognize patterns regardless of their position within the time series data enabling the recognition and detection of events that occur at different timestamps. 
Formally, the time series encoder is defined as follow:
\begin{equation}
TimeSeriesEncoder = M_e\circ C^2_e \circ C^1_e
\end{equation}
where $C^1_e$ and $C^2_e$ are the 1D CNNs with down-sampling. Relu activation is used for each layer. $M_e$ is the encoder's mapping linear layer that projects the feature maps into the same dimension of the text embeddings to obtain the time series embeddings $<time\_series\_embedding>   \in \mathbb{R}^{f \times d}$, where $f$ is the size of the compressed time series, and $d$ is the dimension of the embedding. On the other hand, the time series decoder is responsible for the up-sampling and deconvolution, and is defined as follow:
\begin{equation}
TimeSeriesDecoder = C^2_d \circ C^1_d \circ M_d
\end{equation}
where $C^1_d$ and $C^2_d$ are the 1D deconvolution layers with up-sampling. Relu activation is used for $C^1_d$ and sigmoid activation is used for $C^2_d$. $M_d$ is the decoder's mapping linear layer that projects back the feature maps. Only the time series part of  \textit{TS-Caption Dataset} and \textit{Generated Data} are used to train our 1D CNN autoencoder as it is an unsupervised training and labels (in this case captions) are not used. Given a time series $T=t_1t_2\ldots t_l$, the reconstructed time series from the 1D CNN autoencoder is denoted by $T^r=t_1^rt_2^r\ldots t_l^r$, where $t_i^r,i=1,2,\ldots,l$ is the reconstructed value at the timestamp $i$. The autoencoder is trained by minimizing the $L_1$ loss function between $T$ and $T^r$.
After training, we only need the time series encoder which is frozen and used to extract the time series embeddings that are incorporated into the remaining steps of TSLM training.

\subsection{Denoise Generated Data via Cross-Modal Dense Retrieval Scoring}

Although LLaMA2-13B-Chat can generate promising synthetic data, it may also produce plausible but incorrect factual information, a phenomenon known as hallucinations in LLMs. Therefore, to further improve the quality of synthetic data, we propose a new denoising model via cross-modal dense retrieval scoring. 
As shown in Figure \ref{training_phase} (c), we leverage the groundtruth \textit{TS-Caption Dataset} to train a cross-modal dense retrieval model where the objective is to project the embeddings of each modality into a joint embedding space by capturing the semantic relationships between different modalities. In our case, the two modalities are the joint representation of the time series and the textual representation of the caption. By creating a joint embedding space, we enhance the understanding of the relationship between modalities, and we ensure that the representations are coherent and consistent which enable us to filter noisy time series-caption pairs from the generated data. 
Therefore, 
Once the joint embedding space is learned, the similarity between modalities can be assessed, and pairs with low computed similarity from the generated data can be flagged as noisy, and therefore removed to obtain a \textit{Denoised Generated Data} as shown in Figure \ref{training_phase}.

Formally, given a time series-caption pair $(T,c)$, the Multi-Modal Encoder (Vector) variant is used to extract the vector-based embedding $\boldsymbol{X}_{[CLS]} \in  \mathbb{R}^d$ of the joint representation $JR(T)$ of $T$. For the caption $c$, we add the $[CLS]$ token to the beginning of the sequence and we feed the resulting sequence to the embedding layer and transformer blocks of the text encoder. Then, we pool the hidden state of the $[CLS]$ token from the last transformer block to extract the caption embedding $\boldsymbol{C}_{[CLS]} \in  \mathbb{R}^d$. 
The multi-modal encoder (vector) and the text encoder share the layers $embed\_tokens$ and $transformer\_blocks$. The multi-modal encoder (vector) has an additional reprogramming layer to align the time series embeddings with the text embeddings of the time series as explained previously. As shown in Figure \ref{training_phase} (c), the goal is to create an embedding space such that relevant pairs of time series and captions will have higher similarity than the irrelevant ones. To train the cross-modal dense retrieval model, we need to create positive and negative time series-caption pairs. For a given time series-caption pair $(T,c)$, we denote the negative captions by $c_1^-,c_2^-,\ldots,c_{ng}^-$, where $ng$ is the number of negatives. We update the parameters of the 
dense retrieval model 
by minimizing the log likelihood loss of the relevant caption:
\begin{equation}
\begin{aligned}
& L\left(T, c, c_1^-, \cdots, c_{ng}^-\right)
= & -\log \frac{e^{\operatorname{sim}\left(T, c\right)}}{e^{\operatorname{sim}\left(T, c\right)}+\sum_{j=1}^{ng} e^{\operatorname{sim}\left(T, c_j^-\right)}}
\end{aligned}
\end{equation}
where $\operatorname{sim}\left(T, c\right)$ denotes the similarity between the time series and the caption using the dot product of their vectors:
\begin{equation}
\operatorname{sim}(T, c)=\boldsymbol{X}_{[CLS]}^{\top} \boldsymbol{C}_{[CLS]}
\label{denoising_similarity}
\end{equation}
Given a batch that is composed of $B$ time series-caption pairs, in-batch negatives \cite{KarpukhinOMLWEC20} are used to train the cross-modal dense retrieval model. Let $\boldsymbol{T}_B$ and $\boldsymbol{C}_B$ be the embeddings of time series and captions with dimension $(B\times d)$, respectively, in the batch of size $B$. We compute the similarity scores matrix $\boldsymbol{SIM} = \boldsymbol{X}_B \boldsymbol{C}_B^T \in \mathbb{R}^{B \times B}$, where each row corresponds to a time series with $B$ candidate captions. The caption in the diagonal position represents the groundtruth caption and the remaining $B-1$ captions are negatives.

\begin{figure*}[ht!]
\centering
\includegraphics[width=1\textwidth]{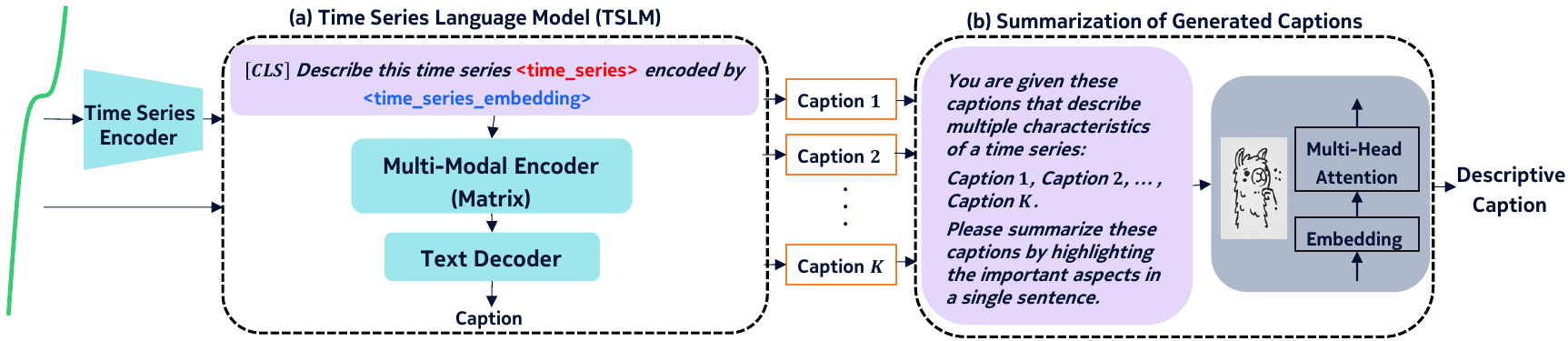}
\caption{The overview of generating a descriptive caption. The joint representation of the unseen time series is extracted by combining the textual and embedding representations, then TSLM generates $K$ captions that are summarized using LLaMA2-13B-Chat to obtain the final descriptive caption. }
\label{testing_phase}
\end{figure*}

After training the cross-modal dense retrieval model using only the groundtruth \textit{TS-Caption Dataset}, we score each pair in the generated data using $\operatorname{sim}\left(T, c\right)$ where high quality generated time series-caption pairs should have larger similarity than low quality generated pairs. To filter the generated data, we remove every pair that has a similarity $\operatorname{sim}\left(T, c\right)$ less than a threshold $Th$. After filtering the generated data, we obtain a \textit{Denoised Generated Data} which is considered as a large-scale and high-quality data for training an accurate time series language model for captioning. Therefore, our new denoising method utilizing the cross-modal dense retrieval scoring enables the use of less effective open-source LLMs, such as LLaMA2-13B-Chat, for synthetic data generation instead of ChatGPT. This approach reduces noise in the generated data and leverages the cost and computational efficiency of open-source LLMs.
\subsection{Time Series Language Model}
The groundtruth \textit{TS-Caption Dataset} and the \textit{Denoised Generated Data} are used to train the TSLM as shown in Figure \ref{training_phase} (d). The Multi-Modal Encoder (Matrix) variant is used in TSLM to encode the joint representation of the time series $T$ and obtain the embedding matrix $\boldsymbol{X} \in  \mathbb{R}^{(n+f) \times d}$.  
The text decoder shares the embedding layer $embed\_tokens$ with the multi-modal encoder, and has decoder transformer blocks and a language model head. Each decoder transformer block is composed of three components which are the self-attention head, cross-attention head, and feed forward layers. The embeddings obtained from the self-attention head depend solely on the generated tokens up to timestamp $t$. To integrate the embeddings $\boldsymbol{X}$ 
into the decoding process, we leverage the cross-attention head which computes embeddings that are both context- and encoder-aware. At each timestamp $t$, the text decoder is expected to output the token corresponding to position $t+1$. The language model head takes the embedding of the sequence of length $t$ from the final decoder block and outputs a probability distribution over the entire vocabulary of the decoder (same vocabulary as the multi-modal encoder). The TSLM is trained using the next token prediction task with the teacher forcing on the caption. 

\section{Generating Descriptive Caption}

After training, TSLM receives an unseen time series as shown in Figure \ref{testing_phase}. 
The joint representation of the unseen time series is composed using the tagged text representation and the time series embeddings that are extracted from the frozen time series encoder. 
We generate $K$ captions, as shown in Figure \ref{testing_phase} (a), that describe multiple aspects of the input time series with a hybrid approach that combines top-p \cite{fan_acl} and top-k \cite{Holtzman2020The} sampling.
To obtain a single-sentence descriptive caption, we query LLaMA2-13B-Chat to summarize the $K$ captions as shown in Figure \ref{testing_phase} (b). 
The used prompt starts with \textit{You are given these captions that describe multiple characteristics of a time series}. Then, we add the $K$ generated captions to the prompt. Finally, we ask LLaMA2-13B-Chat to summarize these captions by adding this part to the prompt: \textit{Please summarize these captions by highlighting the important aspects in a single sentence
}. As output, we obtain the descriptive caption. TSLM acts as a bridge between time series data and LLaMA2-13B-Chat as it generates multiple captions of a time series input that are summarized by the LLaMA2-13B-Chat LLM. This means that the LLM does not need to directly consume the time series modality, which is completely new to the LLM, and it counts on TSLM to translate the time series into multiple textual captions that are understandable by the LLM. The descriptive caption is only evaluated qualitatively in Section \ref{examples_generated_captions}.

\section{Evaluation} \label{eval}

\subsection{Data Collections}

\subsubsection{STOCK Dataset} 
This time series captioning dataset \cite{jhamtani2021truth} is composed of time series data in the form of stock prices that are collected from Google Finance API related to 7 randomly chosen technology companies over a period of 20 years. The dataset contains a total of 1900 time series, each of which consists of a sequence of values. These values are between 0 and 100. There are 3 natural language annotations for each of the 1900 time series leading to a total of 5700 time series-caption pairs. 
Each time series is labeled by three annotators. The annotations can be grouped into these major buckets: trend (increase/decrease trends: 48\%), superlative (max/min values; peaks and troughs: 20\%), volatility (flat/smooth; irregular: 12\%), comparisons (comparison of start and end values: 10\%), and other patterns (10\%). The human annotations are restricted to a maximum of 9 words, so that one annotation focuses only on one pattern of a given time series. 

\subsubsection{Synthetic Time Series (SYNTH)} 
 This time series captioning dataset \cite{jhamtani2021truth} is constructed so that each time series exhibits one of the following six patterns: increases at the beginning, increases in the middle, increases at the end, decreases at the beginning, decreases in the middle, or decreases at the end. The resulting dataset contains a total of 560 time series. There are 3 natural language annotations for each of the 560 time series leading to a total of 1680 time series-caption pairs. 
 Each synthetic time series is generated as follows: first, a trend (increase or decrease) is chosen. This trend is represented by a straight line of length $\leq l/3$, with a random intercept and slope that are chosen within a range based on the selected trend. Next, one of three temporal locations (beginning, middle, end) is randomly picked and considered for the place where the pattern is located within the series, specifically in the first 40 percentile, 30-70 percentile, or 60-100 percentile of the entire length $l$, respectively. The regions that are located outside of the trend remain flat. Finally, more variability is introduced by adding a small noise to each data point. This synthetic setup ensures that the resulting values of the time series are always within the range of (0, 100).

 \textbf{We note that these data scarce time series captioning collections highlight the effectiveness of our new synthetic data generation and denoising steps. We hypothesize that larger datasets would further improve the generalization of TSLM.}

\subsection{Baselines}

\subsubsection{Time Series Captioning}
This group of baselines treat time series as a sequence of numerical values which correspond to the time series modality.

\textbf{TRUCE} \cite{jhamtani2021truth}: This baseline represents a neural truth-conditional model for time series captioning that learns to identify patterns which are relevant to a time series. Three  simple types of modules: \textit{pattern}, \textit{locate}, and \textit{combine} are defined, and then composed together to form handcrafted programs. If a program returns true on the input, the caption generation is conditioned on only the logical program to generate an output text that describes this pattern via a decoder.

\textbf{TSLM (TimeSeries)}: This baseline is a variation of our method that only considers the reprogrammed time series embeddings from the encoder when generating captions 
instead of the joint representations of TSLM.

\textbf{TSLM (w/o denoising)}: This baseline is a variation of our method that uses all the generated data without denoising to train TSLM.

\textbf{TSLM (small)}: This baseline is a variation of our method with a small number of parameters (142 million) compared to the vanilla TSLM (1 billion of parameters).

\textbf{TSLM (medium)}: This baseline is a variation of our method with a medium number of parameters (393 million) compared to the vanilla TSLM (1 billion of parameters).

\subsubsection{Image Captioning}
This group of baselines treat time series as charts or plots which correspond to the image modality. 

\textbf{LLaVA}\footnote{\label{note5}https://huggingface.co/llava-hf/llava-1.5-7b-hf}
: This baseline is a large multi-modal model that leverages a vision encoder and an LLM for general-purpose 
visual and language understanding. LLaVA contains 7 billion parameters  and is fine-tuned with QLoRA \cite{qlora} using the image modality of the time series (chart) for the caption generation task. 

\textbf{LLaVA (ICL)}
: This baseline is a vanilla LLaVA model with in-context learning (ICL) for the caption generation task.


\textbf{PureT} \cite{WangXS22}: This baseline adopts the encoder-decoder framework, where the encoder comprises a SwinTransformer \cite{swintransformer} backbone and multiple refining encoder blocks, while the decoder consists of several decoder blocks. The encoder's role is to extract grid features from the input image and refine them by capturing the intra-relationships between these features. The decoder then generates captions word by word, utilizing the refined image grid features and capturing the inter-relationships between the words and image grid features.

\subsubsection{Text Decoder-Only}
This group of baselines treat time series as a 
sequence of tokens which corresponds to the text modality, and the models of this group are composed of decoders only.

\textbf{LLaMA2-7B-Chat}\footnote{\label{note1}https://huggingface.co/meta-llama/Llama-2-7b-chat-hf}
: This baseline is a decoder-only LLM with 7 billion parameters that is fine-tuned with QLoRA using the textual modality of the time series for the caption generation task.

\textbf{LLaMA2-13B-Chat}\footnote{\label{note2}https://huggingface.co/meta-llama/Llama-2-13b-chat-hf}
: This baseline is a decoder-only LLM with 13 billion parameters that is fine-tuned similar to LLaMA2-7B-Chat.

\textbf{LLaMA2-70B-Chat}\footnote{\label{note4}https://huggingface.co/meta-llama/Llama-2-70b-chat-hf}
: This baseline is a decoder-only LLM with 70 billion parameters that is fine-tuned similar to LLaMA2-7B-Chat.

\textbf{LLaMA3-8B (ICL)}\footnote{\label{note3}https://huggingface.co/meta-llama/Meta-Llama-3-8B-Instruct}
: This baseline is a decoder-only LLM with 8 billion parameters using in-context learning (ICL) for the caption generation task.

\subsubsection{Text Encoder-Decoder}
This group of baselines treat time series as a text or sequence of tokens which corresponds to the text modality, and the models are composed of encoders and decoders.

\textbf{T5} \cite{RaffelSRLNMZLL20}: This baseline is a unified sequence-to-sequence Transformer model that 
standardizes various NLP tasks by converting them into a text-to-text generation format. The encoder processes the input text, and the decoder generates the output text.

\textbf{BART} \cite{LewisLGGMLSZ20}: This baseline utilizes a sequence-to-sequence Transformer architecture with 
denoising pre-training objectives, specifically designed 
to enhance its effectiveness for text generation tasks. Similar to T5, the encoder processes the input text bidirectionally, while the decoder generates output text autoregressively.

\textbf{TSLM (Text)}: This baseline is a variation of our method that only considers the text embeddings from the encoder when generating captions 
instead of the joint representations of TSLM. Compared to the text-based baselines, this variation explicitly injects the positional information into the text-based sequence of the time series with adding the three-phase tags as described in Section \ref{TR}.

\subsection{Experimental Setup}

The time series encoding results in a compressed time series embeddings $f$ with size 6. The dimension of the text and time series embeddings is set to 1024. The $embed\_tokens$ and $transformer\_blocks$ layers are initialized from t5-large\footnote{https://huggingface.co/google-t5/t5-large}
, therefore the vocabulary size $|V|$ is 32,128 and we set the number of text prototypes $p$ to 1000 ($p<<|V|$) as in \cite{jin2024timellm}. For the time series 1D CNN autoencoder, the number of 1D CNN filters for $C^1_e$, $C^2_e$, $C^1_d$, and $C^2_d$ is 512, 256, 256, and 512, respectively. The kernel size is 3, and the stride is 2 for all 1D CNN layers. The dimension of the encoder's mapping linear layer $M_e$ and the decoder's mapping linear layer $M_d$ is $256 \times 1024$ and $1024 \times 256$, respectively. The total number of parameters of TSLM is around 1 billion. 

We apply the in-context prompting data generation for the merged dataset that contains both the training sets of STOCK and SYNTH. Each time series length in the merged dataset varies between 12 and 24. The synthetically generated data has more varieties of sequence lengths ranging from 12 to 50. As explained in Section \ref{data generation}, diversity and quality of the generated data are fundamental criteria in synthetic data generation. To increase the diversity of the generated data, we decided to bootstrap the data generation process by including the already generated data into the groundtruth time series-caption pairs. \textbf{This decision is evaluated based on an intermediate criterion, that reflects the novelty of the generated data, which is the percentage of duplicate samples in the generated data. With bootstrapping, there is only around 2\% of duplicate samples in the generated data, and without bootstrapping, there is around 11\% of duplicate samples in the generated data.} Therefore, it is clear that bootstrapping is very helpful in terms of increasing the diversity of the generated data.
 To improve the quality of data generation, we group the demonstrations by computing the string similarity 
among the annotations.\textbf{ This decision is evaluated based on an intermediate criterion, that reflects the quality of the generated data, which is the percentage of the noisy generated data. With demonstration grouping, the noise level in the generated data is around 7.6\% (equivalent to removing 15,473 pairs from the generated data in the denoising phase), and without demonstration grouping, the noise level is around 12.7\% (equivalent to removing 25,853 pairs from the generated data in the denoising phase).} Therefore, we opt for grouping of demonstrations during the synthetic data generation phase. This grouping makes the number of demonstrations $P$ variable with a maximum of 16. The number of generated samples $S$ in a single LLM pass is equal to 3. The temperature of LLaMA2-13B-Chat for data generation is 0.7. We generated a total number of 203,554 time series-caption pairs. 

We denoise the generated data using our cross-modal dense retrieval model that is trained with a batch size $B$ of 8, and we set the denoising threshold $Th$ to 0, which means that we remove a generated pair that has a negative similarity score computed using Equation (\ref{denoising_similarity}). The denoising step is only applied one time when all the data is generated as opposed to at every round of bootstrap data generation. Here there is a tradeoff between effectiveness and efficiency. Including the denoising step after each single generation during the bootstrap generation significantly increases the synthetic data generation time, and this becomes not practical. So, we can sacrifice some noisy generations in the favor of speeding up the overall process, and at the end we remove all the noisy samples at once. This denoising step results in removing 15,473 generated pairs, and this corresponds to 7.6\% of the total generated data. 
The final size of the denoised generated data is 188,081. 

We use the same training, validation, and testing splits of TRUCE. \textbf{The denoised generated data does not contain any time series-caption pair from the testing sets}. We report the results of TRUCE from their paper \cite{jhamtani2021truth}. For the remaining baselines and TSLM, we use both the original datasets and the denoised generated data for training, then we report the testing results for STOCK and SYNTH. This means that a single model is trained from both datasets. For training TSLM and the baselines, the batch size is 8 and the number of epochs is 10.  For training the 1D CNN autoencoder, the batch size is 32 and the number of epochs is 500. The 1D CNN autoencoder is trained using all the 203,554 generated time series in addition to the original time series from both STOCK and SYNTH. For all training, the optimizer is AdamW \cite{adamw} with a learning rate of 1e-4, a warmup ratio of 0.33, and a linear decay.   
For captions generation, the number of captions $K$ is equal to 3. The top-k is 50 and top-p is 0.95. Some limitations and future directions of TSLM are discussed in Appendix \ref{limitations}.

\subsection{Experimental Results}

We evaluate the performance of TSLM and baselines using three ROUGE \cite{rouge_score} scores: ROUGE-1 (R-1), ROUGE-2 (R-2), and ROUGE-L (R-L); and BERTScore \cite{bertscore}. In addition, we use our trained denoising cross-modal dense retrieval model to report a new score denoted by \textbf{\textit{TSLMScore}}. Given an unseen time series $T$ and a predicted caption $c$, we compute their similarity using the dot product of their embeddings that are extracted using the trained cross-modal dense retrieval model as shown by Equation (\ref{denoising_similarity}). In general, the larger the similarity is, the better the prediction is, as in BERTScore.

\textbf{The generated captions from TSLM and baselines are compared to the groundtruth captions for reporting the quantitative metrics. The descriptive caption that is obtained by summarizing the generated captions from TSLM is only evaluated qualitatively to demonstrate the effectiveness of TSLM in acting as a bridge between the time series data and LLMs.}

\subsubsection{Results on STOCK Dataset}
Table \ref{metrics}(a) shows the performance of different approaches on the STOCK dataset.  We show that our proposed method TSLM outperforms the baselines for all evaluation metrics. By incorporating the textual and embedding representations of a time series into TSLM, we capture both coarse- and fine-grained information of the time series pattern and variation. This leads to a significant improvement over the time series captioning baseline TRUCE both in terms of R-L and BERTScore. The text decoder-only baselines achieve better results than the text encoder-decoder baselines. Fine-tuning LLaMA2 models lead to slightly better results compared to using in-context learning with LLaMA3-8B. However, only using the text modality in the text-based baselines is shown to be less effective than combining both the text and time series modalities in TSLM. The results on this dataset show a clear advantage of our new multi-modal encoder, that fuses both the tagged textual time series and the reprogrammed time series embeddings, in terms of both effectiveness of multiple evaluation metrics and efficiency where TSLM has only 1 billion parameters which is a lot smaller than the LLaMA variations. The efficiency of TSLM translates into a significant reduction in terms of time and memory complexity in both the training and inference phases compared to the LLaMA models. The image modality that is used in fine-tuning LLaVA has a lower performance compared to the text modality used in LLaMA2-70B-Chat, and the combined text and time series modalities used in TSLM. 

TSLMScore reflects the ability to learn intricate variations and patterns better than ROUGE scores and BERTScore as the cross-modal dense retrieval is trained specifically with time series captioning data. TSLMScore shows the advantage of the text- and time series-based models compared to the image-based models. LLaVA (ICL) is in particular not adequate for the time series captioning, as the intricate variations and patterns cannot be captured by the in-context image-based demonstrations, and therefore the generated captions are not accurate.

\begin{table*}[t!]
\small
\caption{ Time series captioning results.
}
\begin{subtable}[t]{0.48\textwidth}
\begin{tabular}{@{}lccccc@{}}
\toprule
\bf Method Name & R-1 & R-2 & R-L &BERTScore&TSLMScore  \\ \midrule
 TRUCE \cite{jhamtani2021truth}& --  &  -- & 50.00 &0.57 & -- \\
 PureT \cite{WangXS22}&57.45  & 37.86 &56.98 &0.73 & 2.82 \\
 LLaVA&60.07 & 41.37 &59.78 & 0.76&4.05 \\
 LLaVA (ICL)&50.00 & 30.00 &48.52 &0.69 &2.68 \\
 LLaMA2-7B-Chat& 60.59 &41.16 & 59.10&0.75 &4.12\\
 LLaMA2-13B-Chat& 60.75 &41.27 &59.91 &0.76 &4.23\\
 LLaMA2-70B-Chat&63.09 &45.65 &63.25 &0.78 &4.32\\
 LLaMA3-8B (ICL)&60.02  & 42.85&59.30 &0.76 &4.32\\
 T5 \cite{RaffelSRLNMZLL20}& 52.34 & 31.69&51.63 & 0.71&3.90\\
 BART \cite{LewisLGGMLSZ20}& 55.90 & 34.79 & 54.72 & 0.72&3.97 \\\bottomrule
TSLM (TimeSeries)   & 65.26  &47.73 &64.57 &0.79&4.30\\
TSLM (Text)   & 64.13  &46.17 & 63.61 &0.77&4.19 \\
TSLM (w/o denoising)   &  62.99 & 44.34&62.27 &0.76&4.20 \\
TSLM (small)   & 65.08  & 47.44& 64.64&0.78&4.38 \\
TSLM (medium)   & 65.51  &48.37 & 64.75&0.79&4.38 \\\bottomrule
TSLM&\textbf{66.74}  &\textbf{49.44}&\textbf{66.45}&\textbf{0.80} &\textbf{4.42}
 \\ \bottomrule
\end{tabular}
\caption{STOCK}
\label{tab:table1_a}
\end{subtable}
\hspace{\fill}
\begin{subtable}[t]{0.48\textwidth}
\begin{tabular}{@{}lccccc@{}}
\toprule
\bf Method Name & R-1 & R-2 & R-L &BERTScore&TSLMScore  \\ \midrule
 TRUCE \cite{jhamtani2021truth}& --  & --  & 74.00 &0.77 &--  \\
 PureT \cite{WangXS22}&74.01  &58.43 &73.42&0.83&5.39  \\
 LLaVA&81.27 &64.46  &80.28 &\textbf{0.88} & 6.82\\
 LLaVA (ICL)&66.19 &51.86  &65.83 &0.79 &5.17 \\
 LLaMA2-7B-Chat& 81.88 &66.61 &80.33 &0.87 &6.80\\
 LLaMA2-13B-Chat& 81.40 &66.77 &80.56 &\textbf{0.88} &6.82\\
 LLaMA2-70B-Chat&82.30 &68.56 &81.02 &\textbf{0.88} &6.85\\
 LLaMA3-8B (ICL)& 79.46 &66.12 &79.03 &0.87 &6.73\\
 T5 \cite{RaffelSRLNMZLL20}& 76.65 &61.26 &75.85 & 0.84&6.75\\
 BART \cite{LewisLGGMLSZ20}& 78.22& 64.66&76.20 &0.85&6.71 \\\bottomrule
TSLM (TimeSeries)   & 81.36  &67.09 &79.45 &\textbf{0.88}&6.95\\
TSLM (Text)   & 82.30  &66.53 &80.40 &0.87&6.81 \\
TSLM (w/o denoising)   & 80.52  & 65.95&77.57 &0.85&6.63 \\
TSLM (small)   & 82.44  &\textbf{72.67} &81.82 &\textbf{0.88}&6.79 \\
TSLM (medium)   & 83.35  &72.43 & 82.40&\textbf{0.88}&6.89 \\\bottomrule
TSLM&\textbf{85.46}  &71.43  &\textbf{83.20} &\textbf{0.88}&\textbf{6.98}
 \\ \bottomrule
\end{tabular}
\caption{SYNTH}
\label{tab:table1_b}
\end{subtable}

\label{metrics}
\end{table*}

\subsubsection{Results on SYNTH Dataset}
Table \ref{metrics}(b) shows the performance of different approaches on the SYNTH dataset. Consistent with STOCK dataset, our results on the SYNTH dataset show the importance of the textual and embedding representations of a time series in improving the time series captioning, and by consequence, TSLM outperforms the baselines for all evaluation metrics. TRUCE uses an LSTM decoder in an auto-regressive manner to generate the caption by considering the embedding of the previous tokens and the input program representation. The results show that our method and the baselines that use transformer-based auto-regressive decoders lead to significantly better evaluation metrics. The image modality used in LLaVA also achieves lower performance than LLaMA2-70B-Chat and TSLM for this dataset demonstrating the effectiveness of the text- and time series-based models compared to the image-based models.

\subsubsection{Ablation Study}
Table \ref{metrics} shows five variations of TSLM. Our ablation study shows the importance of the joint representation of the time series where we combine both coarse- and fine-grained information captured by 
the textual and embedding representations. This joint representation in TSLM outperforms the single-modality variants TSLM (Text) and TSLM (TimeSeries) for all reported metrics. TSLM (Text) differs from the reported text-based baselines by the explicit injection of the positional information into the text-based sequence of the time series with adding the three-phase tags. This is shown to be effective for time series captioning where TSLM (Text) outperforms all text-based baselines (except LLaMA2-70B-Chat because of the huge difference in the model size) by taking advantage of the tagged time series representation. Our ablation study shows that similarly effective results can be obtained from only incorporating the time series embeddings into TSLM (TimeSeries), where subtle variations and intricate patterns are captured by the reprogrammed embeddings leading to accurate generated captions. Denoising the generated data leads to 6.71\% improvement in terms of R-L score compared to TSLM (w/o denoising). This confirms that although LLaMA2-13B-Chat can generate promising synthetic data, it may also produce plausible but incorrect factual information as a result of the hallucinations in LLMs. Therefore, our new denoising method via cross-modal dense retrieval scoring plays an important role in saving TSLM from learning spurious correlations from noisy data. In addition, our denoising method enables the use of less-effective open-source LLMs, such as LLaMA2-13B-Chat, for synthetic data generation instead of ChatGPT while leveraging the cost and computational efficiency of open-source LLMs. 

We also experimented with various sizes of TSLM models to study the effect of the total number of parameters. For TSLM (small), the dimension of the text and time series embedding is 512, therefore,  the time series 1D CNN autoencoder is trained with a dimension of the encoder's mapping linear layer $M_e$ and the decoder's mapping linear layer $M_d$ being set to $256 \times 512$ and $512 \times 256$, respectively. The $embed\_tokens$ and $transformer\_blocks$ layers of TSLM (small) are initialized from t5-small\footnote{https://huggingface.co/google-t5/t5-small}
.  For TSLM (medium), the dimension of the text and time series embedding is 768, therefore,  the time series 1D CNN autoencoder is trained with a dimension of the encoder's mapping linear layer $M_e$ and the decoder's mapping linear layer $M_d$ being set to $256 \times 768$ and $768 \times 256$, respectively. The $embed\_tokens$ and $transformer\_blocks$ layers of TSLM (medium) are initialized from t5-medium\footnote{https://huggingface.co/google-t5/t5-medium}
. Table \ref{metrics} shows that increasing TSLM size leads to better evaluation metrics mainly for R-1, R-L, and TSLMScore. All TSLM model sizes are significantly smaller than the LLaMA baselines while achieving high evaluation metrics which demonstrates the effectiveness of the time series modality that is incorporated into TSLM.

\subsubsection{Examples of Generated Captions}
\label{examples_generated_captions}

\begin{figure*}[t!]
\centering
\includegraphics[width=1\textwidth]{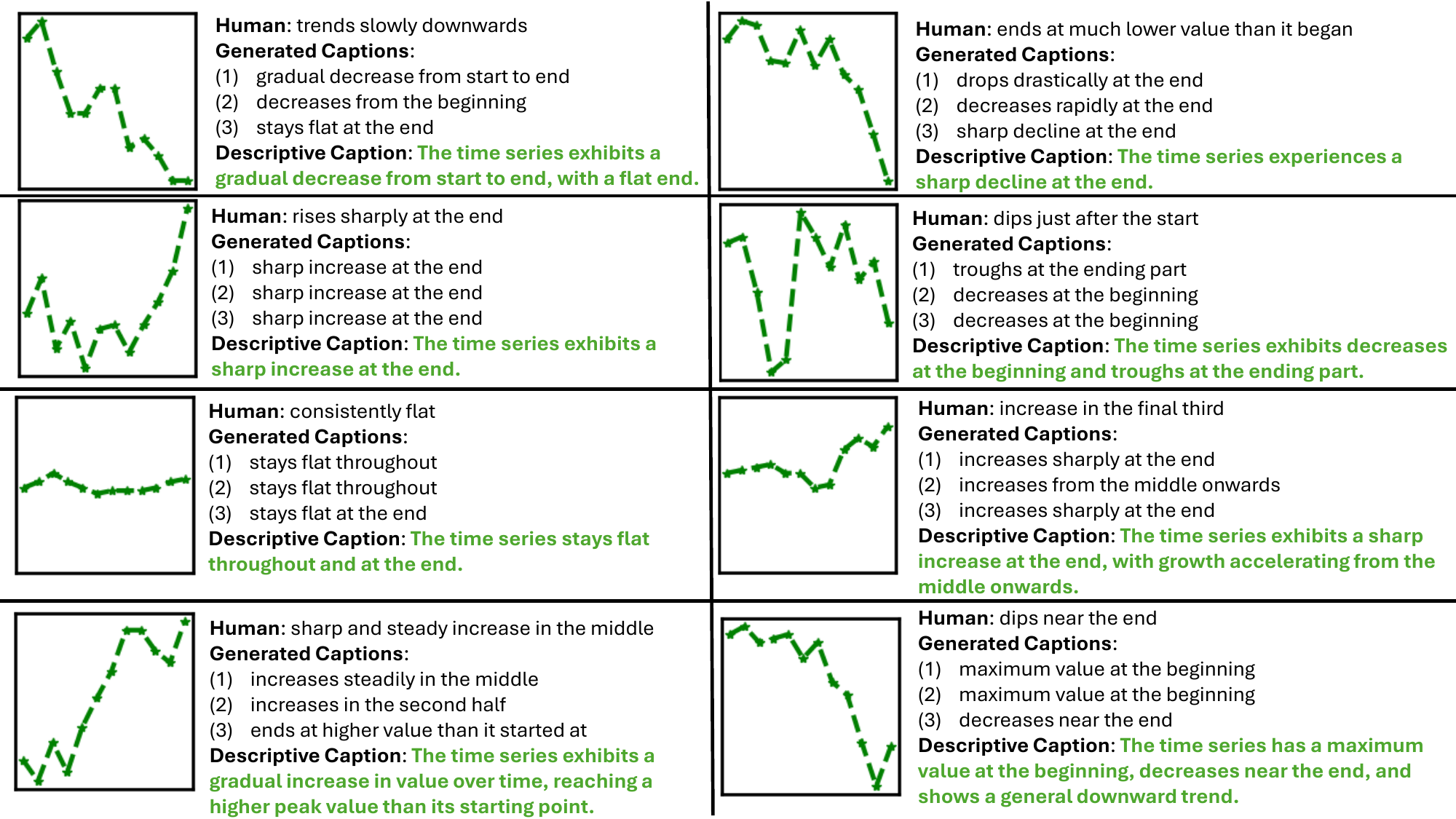}
\caption{STOCK data and generated captions. TSLM generates 3 captions that are summarized using LLaMA2-13B-Chat to obtain a descriptive caption. TSLM generates precise and accurate captions that describe multiple phases and patterns of the time series. }
\label{examples}
\end{figure*}

In Figure \ref{examples}, we show examples of generated captions from TSLM for the STOCK dataset. The caption that corresponds to Human represents the groundtruth caption of the time series. For each case, we show three generated captions from TSLM, and we also show the descriptive caption that results from the summarization of the generated captions using LLaMA2-13B-Chat. The examples show that the generated captions accurately describe multiple phases and patterns of the time series data. By generating multiple captions, TSLM covers most of the important aspects of the time series data. In addition, when a particular pattern is generated by TSLM multiple times, this indicates the high confidence of this specific pattern. The summarization component takes this aspect into account when generating a single descriptive caption from multiple generated captions. Therefore, TSLM acts as a bridge between the time series data and LLaMA2-13B-Chat as it generates multiple captions of a time series input that are summarized by the LLaMA2-13B-Chat LLM. This means that LLaMA2-13B-Chat does not need to directly consume the time series modality, which is completely new to the LLM, and it counts on TSLM to translate the time series into text that is understandable by the LLM. This is an important aspect as recently researchers have focused on allowing LLMs to use external tools \cite{toolformer,toolkengpt,gorilla,toolllm} in order to reduce hallucinations that result from the inability to access up-to-date information, the lack of mathematical skills for precise computations, and  the difficulties in understanding low-resource tasks. The hallucinations that result from applying LLMs directly to time series captioning are primarily related to the lack of mathematical skills of LLMs to precisely describe the patterns of a time series, so any LLM can call TSLM as an external tool to provide multiple precise captions for the time series data.

\section{Conclusions}
In this paper, we proposed a new time series captioning method denoted by TSLM which operates as an encoder-decoder model, leveraging both text prompts and time series data representations to generate precise and accurate textual descriptions of time series inputs. Our model integrates textual- and embedding-based representations of the time series data to capture intricate temporal patterns and variations across multiple phases. We demonstrated the superiority of the time series and text modalities compared to the image modality which is less effective for the time series captioning task as the intricate variations and patterns of the time series are not fully captured by the image-based encoder. 
TSLM acts as a bridge between time series data and LLaMA2-13B-Chat as it generates multiple captions of a time series input that are summarized by the LLaMA2-13B-Chat in order to generate the final descriptive caption. Given the scarcity of time series-caption pairs, TSLM involves a novel data generation that consists of generating synthetic data from existing LLMs to increase the size and diversity of the training dataset, which in turn can improve the robustness and generalization ability of TSLM. The process of synthetically generating data inevitably introduces noisy time series-caption pairs. Therefore, TSLM involves a new denoising method that is applied to the synthetically generated data, before training TSLM with the resulting denoised data. Our experimental results on various time series captioning 
datasets demonstrated that TSLM outperforms existing state-of-the-art approaches from different modalities  by a significant margin. 

Future work includes (1) collecting more time series captioning data from various domains and with longer timestamps to train a domain-agnostic time series language model for time series captioning; and (2) adapting TSLM for the general case of multivariate time series captioning.

\bibliographystyle{ACM-Reference-Format}
\bibliography{main}


\begin{thebibliography}{73}


\ifx \showCODEN    \undefined \def \showCODEN     #1{\unskip}     \fi
\ifx \showDOI      \undefined \def \showDOI       #1{#1}\fi
\ifx \showISBNx    \undefined \def \showISBNx     #1{\unskip}     \fi
\ifx \showISBNxiii \undefined \def \showISBNxiii  #1{\unskip}     \fi
\ifx \showISSN     \undefined \def \showISSN      #1{\unskip}     \fi
\ifx \showLCCN     \undefined \def \showLCCN      #1{\unskip}     \fi
\ifx \shownote     \undefined \def \shownote      #1{#1}          \fi
\ifx \showarticletitle \undefined \def \showarticletitle #1{#1}   \fi
\ifx \showURL      \undefined \def \showURL       {\relax}        \fi
\providecommand\bibfield[2]{#2}
\providecommand\bibinfo[2]{#2}
\providecommand\natexlab[1]{#1}
\providecommand\showeprint[2][]{arXiv:#2}

\bibitem[\protect\citeauthoryear{Alayrac, Donahue, Luc, Miech, Barr, Hasson, Lenc, Mensch, Millican, Reynolds, Ring, Rutherford, Cabi, Han, Gong, Samangooei, Monteiro, Menick, Borgeaud, Brock, Nematzadeh, Sharifzadeh, Binkowski, Barreira, Vinyals, Zisserman, and Simonyan}{Alayrac et~al\mbox{.}}{2022}]%
        {flamingo}
\bibfield{author}{\bibinfo{person}{Jean{-}Baptiste Alayrac}, \bibinfo{person}{Jeff Donahue}, \bibinfo{person}{Pauline Luc}, \bibinfo{person}{Antoine Miech}, \bibinfo{person}{Iain Barr}, \bibinfo{person}{Yana Hasson}, \bibinfo{person}{Karel Lenc}, \bibinfo{person}{Arthur Mensch}, \bibinfo{person}{Katherine Millican}, \bibinfo{person}{Malcolm Reynolds}, \bibinfo{person}{Roman Ring}, \bibinfo{person}{Eliza Rutherford}, \bibinfo{person}{Serkan Cabi}, \bibinfo{person}{Tengda Han}, \bibinfo{person}{Zhitao Gong}, \bibinfo{person}{Sina Samangooei}, \bibinfo{person}{Marianne Monteiro}, \bibinfo{person}{Jacob~L. Menick}, \bibinfo{person}{Sebastian Borgeaud}, \bibinfo{person}{Andy Brock}, \bibinfo{person}{Aida Nematzadeh}, \bibinfo{person}{Sahand Sharifzadeh}, \bibinfo{person}{Mikolaj Binkowski}, \bibinfo{person}{Ricardo Barreira}, \bibinfo{person}{Oriol Vinyals}, \bibinfo{person}{Andrew Zisserman}, {and} \bibinfo{person}{Kar{\'{e}}n Simonyan}.} \bibinfo{year}{2022}\natexlab{}.
\newblock \showarticletitle{Flamingo: a Visual Language Model for Few-Shot Learning}. In \bibinfo{booktitle}{\emph{Advances in Neural Information Processing Systems 35: Annual Conference on Neural Information Processing Systems 2022}}.
\newblock


\bibitem[\protect\citeauthoryear{Andreas, Rohrbach, Darrell, and Klein}{Andreas et~al\mbox{.}}{2016}]%
        {AndreasRDK16}
\bibfield{author}{\bibinfo{person}{Jacob Andreas}, \bibinfo{person}{Marcus Rohrbach}, \bibinfo{person}{Trevor Darrell}, {and} \bibinfo{person}{Dan Klein}.} \bibinfo{year}{2016}\natexlab{}.
\newblock \showarticletitle{Neural Module Networks}. In \bibinfo{booktitle}{\emph{2016 {IEEE} Conference on Computer Vision and Pattern Recognition, {CVPR} 2016}}. \bibinfo{publisher}{{IEEE} Computer Society}, \bibinfo{pages}{39--48}.
\newblock


\bibitem[\protect\citeauthoryear{Berahmand, Daneshfar, Salehi, Li, and Xu}{Berahmand et~al\mbox{.}}{2024}]%
        {BerahmandDSLX24}
\bibfield{author}{\bibinfo{person}{Kamal Berahmand}, \bibinfo{person}{Fatemeh Daneshfar}, \bibinfo{person}{Elaheh~Sadat Salehi}, \bibinfo{person}{Yuefeng Li}, {and} \bibinfo{person}{Yue Xu}.} \bibinfo{year}{2024}\natexlab{}.
\newblock \showarticletitle{Autoencoders and their applications in machine learning: a survey}.
\newblock \bibinfo{journal}{\emph{Artif. Intell. Rev.}} \bibinfo{volume}{57}, \bibinfo{number}{2} (\bibinfo{year}{2024}), \bibinfo{pages}{28}.
\newblock


\bibitem[\protect\citeauthoryear{Chang, Peng, and Chen}{Chang et~al\mbox{.}}{2023}]%
        {chang2024llm4ts}
\bibfield{author}{\bibinfo{person}{Ching Chang}, \bibinfo{person}{Wen{-}Chih Peng}, {and} \bibinfo{person}{Tien{-}Fu Chen}.} \bibinfo{year}{2023}\natexlab{}.
\newblock \showarticletitle{{LLM4TS:} Two-Stage Fine-Tuning for Time-Series Forecasting with Pre-Trained LLMs}.
\newblock \bibinfo{journal}{\emph{CoRR}}  \bibinfo{volume}{abs/2308.08469} (\bibinfo{year}{2023}).
\newblock


\bibitem[\protect\citeauthoryear{Chen, Trabelsi, Heflin, Xu, and Davison}{Chen et~al\mbox{.}}{2020}]%
        {Chen2020TableSU}
\bibfield{author}{\bibinfo{person}{Zhiyu Chen}, \bibinfo{person}{Mohamed Trabelsi}, \bibinfo{person}{Jeff Heflin}, \bibinfo{person}{Yinan Xu}, {and} \bibinfo{person}{Brian~D. Davison}.} \bibinfo{year}{2020}\natexlab{}.
\newblock \showarticletitle{Table Search Using a Deep Contextualized Language Model}. In \bibinfo{booktitle}{\emph{Proceedings of the 43rd International ACM SIGIR Conference on Research and Development in Information Retrieval}}. \bibinfo{publisher}{Association for Computing Machinery}, \bibinfo{address}{New York, NY, USA}, \bibinfo{pages}{589–598}.
\newblock


\bibitem[\protect\citeauthoryear{Cheng, Wen, Liu, and Sun}{Cheng et~al\mbox{.}}{2024}]%
        {abs-2402-02032}
\bibfield{author}{\bibinfo{person}{Hao Cheng}, \bibinfo{person}{Qingsong Wen}, \bibinfo{person}{Yang Liu}, {and} \bibinfo{person}{Liang Sun}.} \bibinfo{year}{2024}\natexlab{}.
\newblock \showarticletitle{RobustTSF: Towards Theory and Design of Robust Time Series Forecasting with Anomalies}.
\newblock \bibinfo{journal}{\emph{CoRR}}  \bibinfo{volume}{abs/2402.02032} (\bibinfo{year}{2024}).
\newblock


\bibitem[\protect\citeauthoryear{Chia, Bing, Poria, and Si}{Chia et~al\mbox{.}}{2022}]%
        {ChiaBPS22}
\bibfield{author}{\bibinfo{person}{Yew~Ken Chia}, \bibinfo{person}{Lidong Bing}, \bibinfo{person}{Soujanya Poria}, {and} \bibinfo{person}{Luo Si}.} \bibinfo{year}{2022}\natexlab{}.
\newblock \showarticletitle{RelationPrompt: Leveraging Prompts to Generate Synthetic Data for Zero-Shot Relation Triplet Extraction}. In \bibinfo{booktitle}{\emph{Findings of the Association for Computational Linguistics: {ACL} 2022, Dublin, Ireland, May 22-27, 2022}}. \bibinfo{publisher}{Association for Computational Linguistics}, \bibinfo{pages}{45--57}.
\newblock


\bibitem[\protect\citeauthoryear{Dai and Callan}{Dai and Callan}{2019}]%
        {dai2019}
\bibfield{author}{\bibinfo{person}{Zhuyun Dai} {and} \bibinfo{person}{Jamie Callan}.} \bibinfo{year}{2019}\natexlab{}.
\newblock \showarticletitle{Deeper Text Understanding for IR with Contextual Neural Language Modeling}. In \bibinfo{booktitle}{\emph{Proceedings of the 42nd International ACM SIGIR Conference on Research and Development in Information Retrieval}}. \bibinfo{numpages}{4}~pages.
\newblock


\bibitem[\protect\citeauthoryear{Dama and Sinoquet}{Dama and Sinoquet}{2021}]%
        {Dama2021TimeSA}
\bibfield{author}{\bibinfo{person}{Fatoumata Dama} {and} \bibinfo{person}{Christine Sinoquet}.} \bibinfo{year}{2021}\natexlab{}.
\newblock \showarticletitle{Time Series Analysis and Modeling to Forecast: a Survey}.
\newblock
\urldef\tempurl%
\url{https://api.semanticscholar.org/CorpusID:237940640}
\showURL{%
\tempurl}


\bibitem[\protect\citeauthoryear{Darban, Webb, Pan, Aggarwal, and Salehi}{Darban et~al\mbox{.}}{2022}]%
        {abs-2211-05244}
\bibfield{author}{\bibinfo{person}{Zahra~Zamanzadeh Darban}, \bibinfo{person}{Geoffrey~I. Webb}, \bibinfo{person}{Shirui Pan}, \bibinfo{person}{Charu~C. Aggarwal}, {and} \bibinfo{person}{Mahsa Salehi}.} \bibinfo{year}{2022}\natexlab{}.
\newblock \showarticletitle{Deep Learning for Time Series Anomaly Detection: {A} Survey}.
\newblock \bibinfo{journal}{\emph{CoRR}}  \bibinfo{volume}{abs/2211.05244} (\bibinfo{year}{2022}).
\newblock


\bibitem[\protect\citeauthoryear{Das, Kong, Sen, and Zhou}{Das et~al\mbox{.}}{2023}]%
        {abs-2310-10688}
\bibfield{author}{\bibinfo{person}{Abhimanyu Das}, \bibinfo{person}{Weihao Kong}, \bibinfo{person}{Rajat Sen}, {and} \bibinfo{person}{Yichen Zhou}.} \bibinfo{year}{2023}\natexlab{}.
\newblock \showarticletitle{A decoder-only foundation model for time-series forecasting}.
\newblock \bibinfo{journal}{\emph{CoRR}}  \bibinfo{volume}{abs/2310.10688} (\bibinfo{year}{2023}).
\newblock


\bibitem[\protect\citeauthoryear{Dettmers, Pagnoni, Holtzman, and Zettlemoyer}{Dettmers et~al\mbox{.}}{2023}]%
        {qlora}
\bibfield{author}{\bibinfo{person}{Tim Dettmers}, \bibinfo{person}{Artidoro Pagnoni}, \bibinfo{person}{Ari Holtzman}, {and} \bibinfo{person}{Luke Zettlemoyer}.} \bibinfo{year}{2023}\natexlab{}.
\newblock \showarticletitle{QLoRA: Efficient Finetuning of Quantized LLMs}. In \bibinfo{booktitle}{\emph{Advances in Neural Information Processing Systems 36: Annual Conference on Neural Information Processing Systems 2023}}.
\newblock


\bibitem[\protect\citeauthoryear{Devlin, Chang, Lee, and Toutanova}{Devlin et~al\mbox{.}}{2019}]%
        {Devlin2019BERTPO}
\bibfield{author}{\bibinfo{person}{Jacob Devlin}, \bibinfo{person}{Ming-Wei Chang}, \bibinfo{person}{Kenton Lee}, {and} \bibinfo{person}{Kristina Toutanova}.} \bibinfo{year}{2019}\natexlab{}.
\newblock \showarticletitle{BERT: Pre-training of Deep Bidirectional Transformers for Language Understanding}. In \bibinfo{booktitle}{\emph{NAACL-HLT}}.
\newblock


\bibitem[\protect\citeauthoryear{Ding, Qin, Liu, Chia, Li, Joty, and Bing}{Ding et~al\mbox{.}}{2023}]%
        {DingQLCLJB23}
\bibfield{author}{\bibinfo{person}{Bosheng Ding}, \bibinfo{person}{Chengwei Qin}, \bibinfo{person}{Linlin Liu}, \bibinfo{person}{Yew~Ken Chia}, \bibinfo{person}{Boyang Li}, \bibinfo{person}{Shafiq Joty}, {and} \bibinfo{person}{Lidong Bing}.} \bibinfo{year}{2023}\natexlab{}.
\newblock \showarticletitle{Is {GPT-3} a Good Data Annotator?}. In \bibinfo{booktitle}{\emph{Proceedings of the 61st Annual Meeting of the Association for Computational Linguistics, {ACL} 2023}}. \bibinfo{publisher}{Association for Computational Linguistics}, \bibinfo{pages}{11173--11195}.
\newblock


\bibitem[\protect\citeauthoryear{Fan, Lewis, and Dauphin}{Fan et~al\mbox{.}}{2018}]%
        {fan_acl}
\bibfield{author}{\bibinfo{person}{Angela Fan}, \bibinfo{person}{Mike Lewis}, {and} \bibinfo{person}{Yann~N. Dauphin}.} \bibinfo{year}{2018}\natexlab{}.
\newblock \showarticletitle{Hierarchical Neural Story Generation}. In \bibinfo{booktitle}{\emph{Proceedings of the 56th Annual Meeting of the Association for Computational Linguistics, {ACL} 2018}}. \bibinfo{publisher}{Association for Computational Linguistics}, \bibinfo{pages}{889--898}.
\newblock


\bibitem[\protect\citeauthoryear{Garza and Canseco}{Garza and Canseco}{2023}]%
        {timegpt}
\bibfield{author}{\bibinfo{person}{Azul Garza} {and} \bibinfo{person}{Max~Mergenthaler Canseco}.} \bibinfo{year}{2023}\natexlab{}.
\newblock \showarticletitle{TimeGPT-1}.
\newblock \bibinfo{journal}{\emph{CoRR}}  \bibinfo{volume}{abs/2310.03589} (\bibinfo{year}{2023}).
\newblock


\bibitem[\protect\citeauthoryear{Gong, Lyu, Zhang, Wang, Zheng, Zhao, Liu, Zhang, Luo, and Chen}{Gong et~al\mbox{.}}{2023}]%
        {multimodal_gpt}
\bibfield{author}{\bibinfo{person}{Tao Gong}, \bibinfo{person}{Chengqi Lyu}, \bibinfo{person}{Shilong Zhang}, \bibinfo{person}{Yudong Wang}, \bibinfo{person}{Miao Zheng}, \bibinfo{person}{Qian Zhao}, \bibinfo{person}{Kuikun Liu}, \bibinfo{person}{Wenwei Zhang}, \bibinfo{person}{Ping Luo}, {and} \bibinfo{person}{Kai Chen}.} \bibinfo{year}{2023}\natexlab{}.
\newblock \showarticletitle{MultiModal-GPT: {A} Vision and Language Model for Dialogue with Humans}.
\newblock \bibinfo{journal}{\emph{CoRR}}  \bibinfo{volume}{abs/2305.04790} (\bibinfo{year}{2023}).
\newblock


\bibitem[\protect\citeauthoryear{Goyal, Li, and Durrett}{Goyal et~al\mbox{.}}{2022}]%
        {abs-2209-12356}
\bibfield{author}{\bibinfo{person}{Tanya Goyal}, \bibinfo{person}{Junyi~Jessy Li}, {and} \bibinfo{person}{Greg Durrett}.} \bibinfo{year}{2022}\natexlab{}.
\newblock \showarticletitle{News Summarization and Evaluation in the Era of {GPT-3}}.
\newblock \bibinfo{journal}{\emph{CoRR}}  \bibinfo{volume}{abs/2209.12356} (\bibinfo{year}{2022}).
\newblock


\bibitem[\protect\citeauthoryear{Han, Yuan, and Trabelsi}{Han et~al\mbox{.}}{2023}]%
        {loggpt}
\bibfield{author}{\bibinfo{person}{Xiao Han}, \bibinfo{person}{Shuhan Yuan}, {and} \bibinfo{person}{Mohamed Trabelsi}.} \bibinfo{year}{2023}\natexlab{}.
\newblock \showarticletitle{LogGPT: Log Anomaly Detection via {GPT}}. In \bibinfo{booktitle}{\emph{{IEEE} International Conference on Big Data, BigData 2023}}. \bibinfo{publisher}{{IEEE}}, \bibinfo{pages}{1117--1122}.
\newblock


\bibitem[\protect\citeauthoryear{Hao, Liu, Wang, and Hu}{Hao et~al\mbox{.}}{2023}]%
        {toolkengpt}
\bibfield{author}{\bibinfo{person}{Shibo Hao}, \bibinfo{person}{Tianyang Liu}, \bibinfo{person}{Zhen Wang}, {and} \bibinfo{person}{Zhiting Hu}.} \bibinfo{year}{2023}\natexlab{}.
\newblock \showarticletitle{ToolkenGPT: Augmenting Frozen Language Models with Massive Tools via Tool Embeddings}. In \bibinfo{booktitle}{\emph{Advances in Neural Information Processing Systems 36: Annual Conference on Neural Information Processing Systems 2023}}.
\newblock


\bibitem[\protect\citeauthoryear{Holtzman, Buys, Du, Forbes, and Choi}{Holtzman et~al\mbox{.}}{2020}]%
        {Holtzman2020The}
\bibfield{author}{\bibinfo{person}{Ari Holtzman}, \bibinfo{person}{Jan Buys}, \bibinfo{person}{Li Du}, \bibinfo{person}{Maxwell Forbes}, {and} \bibinfo{person}{Yejin Choi}.} \bibinfo{year}{2020}\natexlab{}.
\newblock \showarticletitle{The Curious Case of Neural Text Degeneration}. In \bibinfo{booktitle}{\emph{International Conference on Learning Representations}}.
\newblock


\bibitem[\protect\citeauthoryear{Horn, Moor, Bock, Rieck, and Borgwardt}{Horn et~al\mbox{.}}{2020}]%
        {HornMBRB20}
\bibfield{author}{\bibinfo{person}{Max Horn}, \bibinfo{person}{Michael Moor}, \bibinfo{person}{Christian Bock}, \bibinfo{person}{Bastian Rieck}, {and} \bibinfo{person}{Karsten~M. Borgwardt}.} \bibinfo{year}{2020}\natexlab{}.
\newblock \showarticletitle{Set Functions for Time Series}. In \bibinfo{booktitle}{\emph{Proceedings of the 37th International Conference on Machine Learning, {ICML} 2020, 13-18 July 2020, Virtual Event}} \emph{(\bibinfo{series}{Proceedings of Machine Learning Research}, Vol.~\bibinfo{volume}{119})}. \bibinfo{publisher}{{PMLR}}, \bibinfo{pages}{4353--4363}.
\newblock


\bibitem[\protect\citeauthoryear{Huang, Liu, van~der Maaten, and Weinberger}{Huang et~al\mbox{.}}{2017}]%
        {densenet}
\bibfield{author}{\bibinfo{person}{Gao Huang}, \bibinfo{person}{Zhuang Liu}, \bibinfo{person}{Laurens van~der Maaten}, {and} \bibinfo{person}{Kilian~Q. Weinberger}.} \bibinfo{year}{2017}\natexlab{}.
\newblock \showarticletitle{Densely Connected Convolutional Networks}. In \bibinfo{booktitle}{\emph{2017 {IEEE} Conference on Computer Vision and Pattern Recognition, {CVPR} 2017}}. \bibinfo{publisher}{{IEEE} Computer Society}, \bibinfo{pages}{2261--2269}.
\newblock


\bibitem[\protect\citeauthoryear{Huang, Li, Yang, Shi, Chang, Ye, Wu, Hong, Huang, Liu, Ren, Zou, Zhao, and Watanabe}{Huang et~al\mbox{.}}{2024}]%
        {audiogpt}
\bibfield{author}{\bibinfo{person}{Rongjie Huang}, \bibinfo{person}{Mingze Li}, \bibinfo{person}{Dongchao Yang}, \bibinfo{person}{Jiatong Shi}, \bibinfo{person}{Xuankai Chang}, \bibinfo{person}{Zhenhui Ye}, \bibinfo{person}{Yuning Wu}, \bibinfo{person}{Zhiqing Hong}, \bibinfo{person}{Jiawei Huang}, \bibinfo{person}{Jinglin Liu}, \bibinfo{person}{Yi Ren}, \bibinfo{person}{Yuexian Zou}, \bibinfo{person}{Zhou Zhao}, {and} \bibinfo{person}{Shinji Watanabe}.} \bibinfo{year}{2024}\natexlab{}.
\newblock \showarticletitle{AudioGPT: Understanding and Generating Speech, Music, Sound, and Talking Head}. In \bibinfo{booktitle}{\emph{Thirty-Eighth {AAAI} Conference on Artificial Intelligence, {AAAI} 2024, Thirty-Sixth Conference on Innovative Applications of Artificial Intelligence, {IAAI} 2024, Fourteenth Symposium on Educational Advances in Artificial Intelligence, {EAAI} 2014}}. \bibinfo{publisher}{{AAAI} Press}, \bibinfo{pages}{23802--23804}.
\newblock


\bibitem[\protect\citeauthoryear{Jhamtani and Berg-Kirkpatrick}{Jhamtani and Berg-Kirkpatrick}{2021}]%
        {jhamtani2021truth}
\bibfield{author}{\bibinfo{person}{Harsh Jhamtani} {and} \bibinfo{person}{Taylor Berg-Kirkpatrick}.} \bibinfo{year}{2021}\natexlab{}.
\newblock \showarticletitle{Truth-Conditional Captioning of Time Series Data}. In \bibinfo{booktitle}{\emph{EMNLP}}.
\newblock


\bibitem[\protect\citeauthoryear{Jia, Yang, Xia, Chen, Parekh, Pham, Le, Sung, Li, and Duerig}{Jia et~al\mbox{.}}{2021}]%
        {align}
\bibfield{author}{\bibinfo{person}{Chao Jia}, \bibinfo{person}{Yinfei Yang}, \bibinfo{person}{Ye Xia}, \bibinfo{person}{Yi{-}Ting Chen}, \bibinfo{person}{Zarana Parekh}, \bibinfo{person}{Hieu Pham}, \bibinfo{person}{Quoc~V. Le}, \bibinfo{person}{Yun{-}Hsuan Sung}, \bibinfo{person}{Zhen Li}, {and} \bibinfo{person}{Tom Duerig}.} \bibinfo{year}{2021}\natexlab{}.
\newblock \showarticletitle{Scaling Up Visual and Vision-Language Representation Learning With Noisy Text Supervision}. In \bibinfo{booktitle}{\emph{Proceedings of the 38th International Conference on Machine Learning, {ICML} 2021}} \emph{(\bibinfo{series}{Proceedings of Machine Learning Research}, Vol.~\bibinfo{volume}{139})}. \bibinfo{publisher}{{PMLR}}, \bibinfo{pages}{4904--4916}.
\newblock


\bibitem[\protect\citeauthoryear{Jin, Wang, Ma, Chu, Zhang, Shi, Chen, Liang, Li, Pan, and Wen}{Jin et~al\mbox{.}}{2024}]%
        {jin2024timellm}
\bibfield{author}{\bibinfo{person}{Ming Jin}, \bibinfo{person}{Shiyu Wang}, \bibinfo{person}{Lintao Ma}, \bibinfo{person}{Zhixuan Chu}, \bibinfo{person}{James~Y. Zhang}, \bibinfo{person}{Xiaoming Shi}, \bibinfo{person}{Pin-Yu Chen}, \bibinfo{person}{Yuxuan Liang}, \bibinfo{person}{Yuan-Fang Li}, \bibinfo{person}{Shirui Pan}, {and} \bibinfo{person}{Qingsong Wen}.} \bibinfo{year}{2024}\natexlab{}.
\newblock \showarticletitle{Time-{LLM}: Time Series Forecasting by Reprogramming Large Language Models}. In \bibinfo{booktitle}{\emph{The Twelfth International Conference on Learning Representations}}.
\newblock


\bibitem[\protect\citeauthoryear{Kafle, Shrestha, Price, Cohen, and Kanan}{Kafle et~al\mbox{.}}{2020}]%
        {KafleSPCK20}
\bibfield{author}{\bibinfo{person}{Kushal Kafle}, \bibinfo{person}{Robik Shrestha}, \bibinfo{person}{Brian~L. Price}, \bibinfo{person}{Scott Cohen}, {and} \bibinfo{person}{Christopher Kanan}.} \bibinfo{year}{2020}\natexlab{}.
\newblock \showarticletitle{Answering Questions about Data Visualizations using Efficient Bimodal Fusion}. In \bibinfo{booktitle}{\emph{{IEEE} Winter Conference on Applications of Computer Vision, {WACV} 2020}}. \bibinfo{publisher}{{IEEE}}, \bibinfo{pages}{1487--1496}.
\newblock


\bibitem[\protect\citeauthoryear{Karpukhin, Oguz, Min, Lewis, Wu, Edunov, Chen, and Yih}{Karpukhin et~al\mbox{.}}{2020}]%
        {KarpukhinOMLWEC20}
\bibfield{author}{\bibinfo{person}{Vladimir Karpukhin}, \bibinfo{person}{Barlas Oguz}, \bibinfo{person}{Sewon Min}, \bibinfo{person}{Patrick S.~H. Lewis}, \bibinfo{person}{Ledell Wu}, \bibinfo{person}{Sergey Edunov}, \bibinfo{person}{Danqi Chen}, {and} \bibinfo{person}{Wen{-}tau Yih}.} \bibinfo{year}{2020}\natexlab{}.
\newblock \showarticletitle{Dense Passage Retrieval for Open-Domain Question Answering}. In \bibinfo{booktitle}{\emph{Proceedings of the 2020 Conference on Empirical Methods in Natural Language Processing, {EMNLP} 2020}}. \bibinfo{publisher}{Association for Computational Linguistics}, \bibinfo{pages}{6769--6781}.
\newblock


\bibitem[\protect\citeauthoryear{Khattab and Zaharia}{Khattab and Zaharia}{2020}]%
        {khattab_sigir}
\bibfield{author}{\bibinfo{person}{Omar Khattab} {and} \bibinfo{person}{Matei Zaharia}.} \bibinfo{year}{2020}\natexlab{}.
\newblock \showarticletitle{ColBERT: Efficient and Effective Passage Search via Contextualized Late Interaction over {BERT}}. In \bibinfo{booktitle}{\emph{Proceedings of the 43rd International {ACM} {SIGIR} conference on research and development in Information Retrieval, {SIGIR} 2020}}. \bibinfo{publisher}{{ACM}}, \bibinfo{pages}{39--48}.
\newblock


\bibitem[\protect\citeauthoryear{Lewis, Liu, Goyal, Ghazvininejad, Mohamed, Levy, Stoyanov, and Zettlemoyer}{Lewis et~al\mbox{.}}{2020}]%
        {LewisLGGMLSZ20}
\bibfield{author}{\bibinfo{person}{Mike Lewis}, \bibinfo{person}{Yinhan Liu}, \bibinfo{person}{Naman Goyal}, \bibinfo{person}{Marjan Ghazvininejad}, \bibinfo{person}{Abdelrahman Mohamed}, \bibinfo{person}{Omer Levy}, \bibinfo{person}{Veselin Stoyanov}, {and} \bibinfo{person}{Luke Zettlemoyer}.} \bibinfo{year}{2020}\natexlab{}.
\newblock \showarticletitle{{BART:} Denoising Sequence-to-Sequence Pre-training for Natural Language Generation, Translation, and Comprehension}. In \bibinfo{booktitle}{\emph{Proceedings of the 58th Annual Meeting of the Association for Computational Linguistics, {ACL} 2020}}. \bibinfo{publisher}{Association for Computational Linguistics}, \bibinfo{pages}{7871--7880}.
\newblock


\bibitem[\protect\citeauthoryear{Li, Pei, and Li}{Li et~al\mbox{.}}{2023}]%
        {LiPL23}
\bibfield{author}{\bibinfo{person}{Pengzhi Li}, \bibinfo{person}{Yan Pei}, {and} \bibinfo{person}{Jianqiang Li}.} \bibinfo{year}{2023}\natexlab{}.
\newblock \showarticletitle{A comprehensive survey on design and application of autoencoder in deep learning}.
\newblock \bibinfo{journal}{\emph{Appl. Soft Comput.}}  \bibinfo{volume}{138} (\bibinfo{year}{2023}), \bibinfo{pages}{110176}.
\newblock


\bibitem[\protect\citeauthoryear{Liang, Wen, Nie, Jiang, Jin, Song, Pan, and Wen}{Liang et~al\mbox{.}}{2024}]%
        {liang_time_series_analysis_survey}
\bibfield{author}{\bibinfo{person}{Yuxuan Liang}, \bibinfo{person}{Haomin Wen}, \bibinfo{person}{Yuqi Nie}, \bibinfo{person}{Yushan Jiang}, \bibinfo{person}{Ming Jin}, \bibinfo{person}{Dongjin Song}, \bibinfo{person}{Shirui Pan}, {and} \bibinfo{person}{Qingsong Wen}.} \bibinfo{year}{2024}\natexlab{}.
\newblock \showarticletitle{Foundation Models for Time Series Analysis: {A} Tutorial and Survey}.
\newblock \bibinfo{journal}{\emph{CoRR}}  \bibinfo{volume}{abs/2403.14735} (\bibinfo{year}{2024}).
\newblock


\bibitem[\protect\citeauthoryear{Lin}{Lin}{2004}]%
        {rouge_score}
\bibfield{author}{\bibinfo{person}{Chin-Yew Lin}.} \bibinfo{year}{2004}\natexlab{}.
\newblock \showarticletitle{{ROUGE}: A Package for Automatic Evaluation of Summaries}. In \bibinfo{booktitle}{\emph{Text Summarization Branches Out}}. \bibinfo{publisher}{Association for Computational Linguistics}, \bibinfo{address}{Barcelona, Spain}, \bibinfo{pages}{74--81}.
\newblock


\bibitem[\protect\citeauthoryear{Liu, Li, Wu, and Lee}{Liu et~al\mbox{.}}{2023}]%
        {llava}
\bibfield{author}{\bibinfo{person}{Haotian Liu}, \bibinfo{person}{Chunyuan Li}, \bibinfo{person}{Qingyang Wu}, {and} \bibinfo{person}{Yong~Jae Lee}.} \bibinfo{year}{2023}\natexlab{}.
\newblock \showarticletitle{Visual Instruction Tuning}. In \bibinfo{booktitle}{\emph{Advances in Neural Information Processing Systems 36: Annual Conference on Neural Information Processing Systems 2023}}.
\newblock


\bibitem[\protect\citeauthoryear{Liu, Ott, Goyal, Du, Joshi, Chen, Levy, Lewis, Zettlemoyer, and Stoyanov}{Liu et~al\mbox{.}}{2019}]%
        {Liu2019RoBERTaAR}
\bibfield{author}{\bibinfo{person}{Yinhan Liu}, \bibinfo{person}{Myle Ott}, \bibinfo{person}{Naman Goyal}, \bibinfo{person}{Jingfei Du}, \bibinfo{person}{Mandar Joshi}, \bibinfo{person}{Danqi Chen}, \bibinfo{person}{Omer Levy}, \bibinfo{person}{Mike Lewis}, \bibinfo{person}{Luke Zettlemoyer}, {and} \bibinfo{person}{Veselin Stoyanov}.} \bibinfo{year}{2019}\natexlab{}.
\newblock \showarticletitle{RoBERTa: A Robustly Optimized BERT Pretraining Approach}.
\newblock \bibinfo{journal}{\emph{ArXiv}}  \bibinfo{volume}{abs/1907.11692} (\bibinfo{year}{2019}).
\newblock


\bibitem[\protect\citeauthoryear{Liu, Lin, Cao, Hu, Wei, Zhang, Lin, and Guo}{Liu et~al\mbox{.}}{2021}]%
        {swintransformer}
\bibfield{author}{\bibinfo{person}{Ze Liu}, \bibinfo{person}{Yutong Lin}, \bibinfo{person}{Yue Cao}, \bibinfo{person}{Han Hu}, \bibinfo{person}{Yixuan Wei}, \bibinfo{person}{Zheng Zhang}, \bibinfo{person}{Stephen Lin}, {and} \bibinfo{person}{Baining Guo}.} \bibinfo{year}{2021}\natexlab{}.
\newblock \showarticletitle{Swin Transformer: Hierarchical Vision Transformer using Shifted Windows}. In \bibinfo{booktitle}{\emph{2021 {IEEE/CVF} International Conference on Computer Vision, {ICCV} 2021}}. \bibinfo{publisher}{{IEEE}}, \bibinfo{pages}{9992--10002}.
\newblock


\bibitem[\protect\citeauthoryear{Loshchilov and Hutter}{Loshchilov and Hutter}{2019}]%
        {adamw}
\bibfield{author}{\bibinfo{person}{Ilya Loshchilov} {and} \bibinfo{person}{Frank Hutter}.} \bibinfo{year}{2019}\natexlab{}.
\newblock \showarticletitle{Decoupled Weight Decay Regularization}. In \bibinfo{booktitle}{\emph{7th International Conference on Learning Representations, {ICLR} 2019}}. \bibinfo{publisher}{OpenReview.net}.
\newblock


\bibitem[\protect\citeauthoryear{Mahinpei, Kostic, and Tanner}{Mahinpei et~al\mbox{.}}{2022}]%
        {MahinpeiKT22}
\bibfield{author}{\bibinfo{person}{Anita Mahinpei}, \bibinfo{person}{Zona Kostic}, {and} \bibinfo{person}{Chris Tanner}.} \bibinfo{year}{2022}\natexlab{}.
\newblock \showarticletitle{LineCap: Line Charts for Data Visualization Captioning Models}. In \bibinfo{booktitle}{\emph{2022 {IEEE} Visualization and Visual Analytics (VIS)}}. \bibinfo{publisher}{{IEEE}}, \bibinfo{pages}{35--39}.
\newblock


\bibitem[\protect\citeauthoryear{Matsubara, Sakurai, van Panhuis, and Faloutsos}{Matsubara et~al\mbox{.}}{2014}]%
        {MatsubaraSPF14}
\bibfield{author}{\bibinfo{person}{Yasuko Matsubara}, \bibinfo{person}{Yasushi Sakurai}, \bibinfo{person}{Willem~G. van Panhuis}, {and} \bibinfo{person}{Christos Faloutsos}.} \bibinfo{year}{2014}\natexlab{}.
\newblock \showarticletitle{{FUNNEL:} automatic mining of spatially coevolving epidemics}. In \bibinfo{booktitle}{\emph{The 20th {ACM} {SIGKDD} International Conference on Knowledge Discovery and Data Mining, 2014}}. \bibinfo{publisher}{{ACM}}, \bibinfo{pages}{105--114}.
\newblock


\bibitem[\protect\citeauthoryear{Min, Ross, Sulem, Veyseh, Nguyen, Sainz, Agirre, Heintz, and Roth}{Min et~al\mbox{.}}{2024}]%
        {MinRSVNSAHR24}
\bibfield{author}{\bibinfo{person}{Bonan Min}, \bibinfo{person}{Hayley Ross}, \bibinfo{person}{Elior Sulem}, \bibinfo{person}{Amir Pouran~Ben Veyseh}, \bibinfo{person}{Thien~Huu Nguyen}, \bibinfo{person}{Oscar Sainz}, \bibinfo{person}{Eneko Agirre}, \bibinfo{person}{Ilana Heintz}, {and} \bibinfo{person}{Dan Roth}.} \bibinfo{year}{2024}\natexlab{}.
\newblock \showarticletitle{Recent Advances in Natural Language Processing via Large Pre-trained Language Models: {A} Survey}.
\newblock \bibinfo{journal}{\emph{{ACM} Comput. Surv.}} \bibinfo{volume}{56}, \bibinfo{number}{2} (\bibinfo{year}{2024}), \bibinfo{pages}{30:1--30:40}.
\newblock


\bibitem[\protect\citeauthoryear{Murakami, Watanabe, Miyazawa, Goshima, Yanase, Takamura, and Miyao}{Murakami et~al\mbox{.}}{2017}]%
        {MurakamiWMGYTM17}
\bibfield{author}{\bibinfo{person}{Soichiro Murakami}, \bibinfo{person}{Akihiko Watanabe}, \bibinfo{person}{Akira Miyazawa}, \bibinfo{person}{Keiichi Goshima}, \bibinfo{person}{Toshihiko Yanase}, \bibinfo{person}{Hiroya Takamura}, {and} \bibinfo{person}{Yusuke Miyao}.} \bibinfo{year}{2017}\natexlab{}.
\newblock \showarticletitle{Learning to Generate Market Comments from Stock Prices}. In \bibinfo{booktitle}{\emph{Proceedings of the 55th Annual Meeting of the Association for Computational Linguistics, {ACL} 2017}}. \bibinfo{publisher}{Association for Computational Linguistics}, \bibinfo{pages}{1374--1384}.
\newblock


\bibitem[\protect\citeauthoryear{Nogueira and Cho}{Nogueira and Cho}{2019}]%
        {nogueira_passage}
\bibfield{author}{\bibinfo{person}{Rodrigo~Frassetto Nogueira} {and} \bibinfo{person}{Kyunghyun Cho}.} \bibinfo{year}{2019}\natexlab{}.
\newblock \showarticletitle{Passage Re-ranking with {BERT}}.
\newblock \bibinfo{journal}{\emph{CoRR}}  \bibinfo{volume}{abs/1901.04085} (\bibinfo{year}{2019}).
\newblock


\bibitem[\protect\citeauthoryear{Nogueira, Yang, Cho, and Lin}{Nogueira et~al\mbox{.}}{2019}]%
        {nogueira_multi}
\bibfield{author}{\bibinfo{person}{Rodrigo~Frassetto Nogueira}, \bibinfo{person}{Wei Yang}, \bibinfo{person}{Kyunghyun Cho}, {and} \bibinfo{person}{Jimmy Lin}.} \bibinfo{year}{2019}\natexlab{}.
\newblock \showarticletitle{Multi-Stage Document Ranking with {BERT}}.
\newblock \bibinfo{journal}{\emph{CoRR}}  \bibinfo{volume}{abs/1910.14424} (\bibinfo{year}{2019}).
\newblock


\bibitem[\protect\citeauthoryear{Patil, Zhang, Wang, and Gonzalez}{Patil et~al\mbox{.}}{2023}]%
        {gorilla}
\bibfield{author}{\bibinfo{person}{Shishir~G. Patil}, \bibinfo{person}{Tianjun Zhang}, \bibinfo{person}{Xin Wang}, {and} \bibinfo{person}{Joseph~E. Gonzalez}.} \bibinfo{year}{2023}\natexlab{}.
\newblock \showarticletitle{Gorilla: Large Language Model Connected with Massive APIs}.
\newblock \bibinfo{journal}{\emph{CoRR}}  \bibinfo{volume}{abs/2305.15334} (\bibinfo{year}{2023}).
\newblock


\bibitem[\protect\citeauthoryear{Qin, Liang, Ye, Zhu, Yan, Lu, Lin, Cong, Tang, Qian, Zhao, Tian, Xie, Zhou, Gerstein, Li, Liu, and Sun}{Qin et~al\mbox{.}}{2023}]%
        {toolllm}
\bibfield{author}{\bibinfo{person}{Yujia Qin}, \bibinfo{person}{Shihao Liang}, \bibinfo{person}{Yining Ye}, \bibinfo{person}{Kunlun Zhu}, \bibinfo{person}{Lan Yan}, \bibinfo{person}{Yaxi Lu}, \bibinfo{person}{Yankai Lin}, \bibinfo{person}{Xin Cong}, \bibinfo{person}{Xiangru Tang}, \bibinfo{person}{Bill Qian}, \bibinfo{person}{Sihan Zhao}, \bibinfo{person}{Runchu Tian}, \bibinfo{person}{Ruobing Xie}, \bibinfo{person}{Jie Zhou}, \bibinfo{person}{Mark Gerstein}, \bibinfo{person}{Dahai Li}, \bibinfo{person}{Zhiyuan Liu}, {and} \bibinfo{person}{Maosong Sun}.} \bibinfo{year}{2023}\natexlab{}.
\newblock \showarticletitle{ToolLLM: Facilitating Large Language Models to Master 16000+ Real-world APIs}.
\newblock \bibinfo{journal}{\emph{CoRR}}  \bibinfo{volume}{abs/2307.16789} (\bibinfo{year}{2023}).
\newblock


\bibitem[\protect\citeauthoryear{Radford, Kim, Hallacy, Ramesh, Goh, Agarwal, Sastry, Askell, Mishkin, Clark, Krueger, and Sutskever}{Radford et~al\mbox{.}}{2021}]%
        {clip}
\bibfield{author}{\bibinfo{person}{Alec Radford}, \bibinfo{person}{Jong~Wook Kim}, \bibinfo{person}{Chris Hallacy}, \bibinfo{person}{Aditya Ramesh}, \bibinfo{person}{Gabriel Goh}, \bibinfo{person}{Sandhini Agarwal}, \bibinfo{person}{Girish Sastry}, \bibinfo{person}{Amanda Askell}, \bibinfo{person}{Pamela Mishkin}, \bibinfo{person}{Jack Clark}, \bibinfo{person}{Gretchen Krueger}, {and} \bibinfo{person}{Ilya Sutskever}.} \bibinfo{year}{2021}\natexlab{}.
\newblock \showarticletitle{Learning Transferable Visual Models From Natural Language Supervision}. In \bibinfo{booktitle}{\emph{Proceedings of the 38th International Conference on Machine Learning, {ICML} 2021}} \emph{(\bibinfo{series}{Proceedings of Machine Learning Research}, Vol.~\bibinfo{volume}{139})}. \bibinfo{publisher}{{PMLR}}, \bibinfo{pages}{8748--8763}.
\newblock


\bibitem[\protect\citeauthoryear{Raffel, Shazeer, Roberts, Lee, Narang, Matena, Zhou, Li, and Liu}{Raffel et~al\mbox{.}}{2020}]%
        {RaffelSRLNMZLL20}
\bibfield{author}{\bibinfo{person}{Colin Raffel}, \bibinfo{person}{Noam Shazeer}, \bibinfo{person}{Adam Roberts}, \bibinfo{person}{Katherine Lee}, \bibinfo{person}{Sharan Narang}, \bibinfo{person}{Michael Matena}, \bibinfo{person}{Yanqi Zhou}, \bibinfo{person}{Wei Li}, {and} \bibinfo{person}{Peter~J. Liu}.} \bibinfo{year}{2020}\natexlab{}.
\newblock \showarticletitle{Exploring the Limits of Transfer Learning with a Unified Text-to-Text Transformer}.
\newblock \bibinfo{journal}{\emph{J. Mach. Learn. Res.}}  \bibinfo{volume}{21} (\bibinfo{year}{2020}), \bibinfo{pages}{140:1--140:67}.
\newblock


\bibitem[\protect\citeauthoryear{Ramesh, Pavlov, Goh, Gray, Voss, Radford, Chen, and Sutskever}{Ramesh et~al\mbox{.}}{2021}]%
        {dalle}
\bibfield{author}{\bibinfo{person}{Aditya Ramesh}, \bibinfo{person}{Mikhail Pavlov}, \bibinfo{person}{Gabriel Goh}, \bibinfo{person}{Scott Gray}, \bibinfo{person}{Chelsea Voss}, \bibinfo{person}{Alec Radford}, \bibinfo{person}{Mark Chen}, {and} \bibinfo{person}{Ilya Sutskever}.} \bibinfo{year}{2021}\natexlab{}.
\newblock \showarticletitle{Zero-Shot Text-to-Image Generation}. In \bibinfo{booktitle}{\emph{Proceedings of the 38th International Conference on Machine Learning, {ICML} 2021}} \emph{(\bibinfo{series}{Proceedings of Machine Learning Research}, Vol.~\bibinfo{volume}{139})}. \bibinfo{publisher}{{PMLR}}, \bibinfo{pages}{8821--8831}.
\newblock


\bibitem[\protect\citeauthoryear{Sakata, Shibata, Tanaka, and Kurohashi}{Sakata et~al\mbox{.}}{2019}]%
        {sakata2019}
\bibfield{author}{\bibinfo{person}{Wataru Sakata}, \bibinfo{person}{Tomohide Shibata}, \bibinfo{person}{Ribeka Tanaka}, {and} \bibinfo{person}{Sadao Kurohashi}.} \bibinfo{year}{2019}\natexlab{}.
\newblock \showarticletitle{FAQ Retrieval Using Query-Question Similarity and BERT-Based Query-Answer Relevance}. In \bibinfo{booktitle}{\emph{Proceedings of the 42nd International ACM SIGIR Conference on Research and Development in Information Retrieval}}. \bibinfo{publisher}{Association for Computing Machinery}, \bibinfo{pages}{1113–1116}.
\newblock
\showISBNx{9781450361729}


\bibitem[\protect\citeauthoryear{Schick, Dwivedi{-}Yu, Dess{\`{\i}}, Raileanu, Lomeli, Hambro, Zettlemoyer, Cancedda, and Scialom}{Schick et~al\mbox{.}}{2023}]%
        {toolformer}
\bibfield{author}{\bibinfo{person}{Timo Schick}, \bibinfo{person}{Jane Dwivedi{-}Yu}, \bibinfo{person}{Roberto Dess{\`{\i}}}, \bibinfo{person}{Roberta Raileanu}, \bibinfo{person}{Maria Lomeli}, \bibinfo{person}{Eric Hambro}, \bibinfo{person}{Luke Zettlemoyer}, \bibinfo{person}{Nicola Cancedda}, {and} \bibinfo{person}{Thomas Scialom}.} \bibinfo{year}{2023}\natexlab{}.
\newblock \showarticletitle{Toolformer: Language Models Can Teach Themselves to Use Tools}. In \bibinfo{booktitle}{\emph{Advances in Neural Information Processing Systems 36: Annual Conference on Neural Information Processing Systems 2023}}.
\newblock


\bibitem[\protect\citeauthoryear{Sowdaboina, Chakraborti, and Sripada}{Sowdaboina et~al\mbox{.}}{2014}]%
        {SowdaboinaCS14}
\bibfield{author}{\bibinfo{person}{Pranay Kumar~Venkata Sowdaboina}, \bibinfo{person}{Sutanu Chakraborti}, {and} \bibinfo{person}{Somayajulu Sripada}.} \bibinfo{year}{2014}\natexlab{}.
\newblock \showarticletitle{Learning to Summarize Time Series Data}. In \bibinfo{booktitle}{\emph{Computational Linguistics and Intelligent Text Processing - 15th International Conference, CICLing 2014}} \emph{(\bibinfo{series}{Lecture Notes in Computer Science}, Vol.~\bibinfo{volume}{8403})}. \bibinfo{publisher}{Springer}, \bibinfo{pages}{515--528}.
\newblock


\bibitem[\protect\citeauthoryear{Sun, Li, Li, Wu, Guo, Zhang, and Wang}{Sun et~al\mbox{.}}{2023}]%
        {SunL0WGZ023}
\bibfield{author}{\bibinfo{person}{Xiaofei Sun}, \bibinfo{person}{Xiaoya Li}, \bibinfo{person}{Jiwei Li}, \bibinfo{person}{Fei Wu}, \bibinfo{person}{Shangwei Guo}, \bibinfo{person}{Tianwei Zhang}, {and} \bibinfo{person}{Guoyin Wang}.} \bibinfo{year}{2023}\natexlab{}.
\newblock \showarticletitle{Text Classification via Large Language Models}. In \bibinfo{booktitle}{\emph{Findings of the Association for Computational Linguistics: {EMNLP} 2023, Singapore, December 6-10, 2023}}. \bibinfo{publisher}{Association for Computational Linguistics}, \bibinfo{pages}{8990--9005}.
\newblock


\bibitem[\protect\citeauthoryear{Trabelsi, Cao, and Heflin}{Trabelsi et~al\mbox{.}}{2020}]%
        {selab_arxiv}
\bibfield{author}{\bibinfo{person}{Mohamed Trabelsi}, \bibinfo{person}{Jin Cao}, {and} \bibinfo{person}{Jeff Heflin}.} \bibinfo{year}{2020}\natexlab{}.
\newblock \showarticletitle{Semantic Labeling Using a Deep Contextualized Language Model}.
\newblock \bibinfo{journal}{\emph{CoRR}}  \bibinfo{volume}{abs/2010.16037} (\bibinfo{year}{2020}).
\newblock


\bibitem[\protect\citeauthoryear{Trabelsi, Cao, and Heflin}{Trabelsi et~al\mbox{.}}{2021a}]%
        {selab}
\bibfield{author}{\bibinfo{person}{Mohamed Trabelsi}, \bibinfo{person}{Jin Cao}, {and} \bibinfo{person}{Jeff Heflin}.} \bibinfo{year}{2021}\natexlab{a}.
\newblock \showarticletitle{SeLaB: Semantic Labeling with BERT}. In \bibinfo{booktitle}{\emph{2021 International Joint Conference on Neural Networks (IJCNN)}}. \bibinfo{pages}{1--8}.
\newblock
\urldef\tempurl%
\url{https://doi.org/10.1109/IJCNN52387.2021.9534408}
\showDOI{\tempurl}


\bibitem[\protect\citeauthoryear{Trabelsi, Chen, Davison, and Heflin}{Trabelsi et~al\mbox{.}}{2021b}]%
        {survey_doc_retrieval}
\bibfield{author}{\bibinfo{person}{Mohamed Trabelsi}, \bibinfo{person}{Zhiyu Chen}, \bibinfo{person}{Brian~D. Davison}, {and} \bibinfo{person}{Jeff Heflin}.} \bibinfo{year}{2021}\natexlab{b}.
\newblock \showarticletitle{Neural ranking models for document retrieval}.
\newblock \bibinfo{journal}{\emph{Inf. Retr. J.}} \bibinfo{volume}{24}, \bibinfo{number}{6} (\bibinfo{year}{2021}), \bibinfo{pages}{400--444}.
\newblock


\bibitem[\protect\citeauthoryear{Trabelsi, Chen, Zhang, Davison, and Heflin}{Trabelsi et~al\mbox{.}}{2022a}]%
        {strubert}
\bibfield{author}{\bibinfo{person}{Mohamed Trabelsi}, \bibinfo{person}{Zhiyu Chen}, \bibinfo{person}{Shuo Zhang}, \bibinfo{person}{Brian~D. Davison}, {and} \bibinfo{person}{Jeff Heflin}.} \bibinfo{year}{2022}\natexlab{a}.
\newblock \showarticletitle{StruBERT: Structure-aware BERT for Table Search and Matching}. In \bibinfo{booktitle}{\emph{Proceedings of the Web Conference (WWW 2022)}}.
\newblock


\bibitem[\protect\citeauthoryear{Trabelsi, Heflin, and Cao}{Trabelsi et~al\mbox{.}}{2022b}]%
        {dame}
\bibfield{author}{\bibinfo{person}{Mohamed Trabelsi}, \bibinfo{person}{Jeff Heflin}, {and} \bibinfo{person}{Jin Cao}.} \bibinfo{year}{2022}\natexlab{b}.
\newblock \showarticletitle{DAME: Domain Adaptation for Matching Entities}. In \bibinfo{booktitle}{\emph{Proceedings of the 15th ACM International Conference on Web Search and Data Mining (WSDM 2022)}}.
\newblock


\bibitem[\protect\citeauthoryear{Trabelsi and Uzunalioglu}{Trabelsi and Uzunalioglu}{2023}]%
        {absformer}
\bibfield{author}{\bibinfo{person}{Mohamed Trabelsi} {and} \bibinfo{person}{H{\"{u}}seyin Uzunalioglu}.} \bibinfo{year}{2023}\natexlab{}.
\newblock \showarticletitle{Absformer: Transformer-Based Model for Unsupervised Multi-Document Abstractive Summarization}. In \bibinfo{booktitle}{\emph{Document Analysis and Recognition - {ICDAR} 2023 Workshops}} \emph{(\bibinfo{series}{Lecture Notes in Computer Science}, Vol.~\bibinfo{volume}{14194})}. \bibinfo{publisher}{Springer}, \bibinfo{pages}{151--166}.
\newblock


\bibitem[\protect\citeauthoryear{Wadden, Wennberg, Luan, and Hajishirzi}{Wadden et~al\mbox{.}}{2019}]%
        {WaddenWLH19}
\bibfield{author}{\bibinfo{person}{David Wadden}, \bibinfo{person}{Ulme Wennberg}, \bibinfo{person}{Yi Luan}, {and} \bibinfo{person}{Hannaneh Hajishirzi}.} \bibinfo{year}{2019}\natexlab{}.
\newblock \showarticletitle{Entity, Relation, and Event Extraction with Contextualized Span Representations}. In \bibinfo{booktitle}{\emph{Proceedings of the 2019 Conference on Empirical Methods in Natural Language Processing and the 9th International Joint Conference on Natural Language Processing, {EMNLP-IJCNLP} 2019}}. \bibinfo{publisher}{Association for Computational Linguistics}, \bibinfo{pages}{5783--5788}.
\newblock


\bibitem[\protect\citeauthoryear{Wang, Hu, Cao, and Sun}{Wang et~al\mbox{.}}{2020}]%
        {WangHCS20}
\bibfield{author}{\bibinfo{person}{Difeng Wang}, \bibinfo{person}{Wei Hu}, \bibinfo{person}{Ermei Cao}, {and} \bibinfo{person}{Weijian Sun}.} \bibinfo{year}{2020}\natexlab{}.
\newblock \showarticletitle{Global-to-Local Neural Networks for Document-Level Relation Extraction}. In \bibinfo{booktitle}{\emph{Proceedings of the 2020 Conference on Empirical Methods in Natural Language Processing, {EMNLP} 2020}}. \bibinfo{publisher}{Association for Computational Linguistics}, \bibinfo{pages}{3711--3721}.
\newblock


\bibitem[\protect\citeauthoryear{Wang, Yang, and Wei}{Wang et~al\mbox{.}}{2024}]%
        {WangYW24}
\bibfield{author}{\bibinfo{person}{Liang Wang}, \bibinfo{person}{Nan Yang}, {and} \bibinfo{person}{Furu Wei}.} \bibinfo{year}{2024}\natexlab{}.
\newblock \showarticletitle{Learning to Retrieve In-Context Examples for Large Language Models}. In \bibinfo{booktitle}{\emph{Proceedings of the 18th Conference of the European Chapter of the Association for Computational Linguistics}}. \bibinfo{publisher}{Association for Computational Linguistics}, \bibinfo{pages}{1752--1767}.
\newblock


\bibitem[\protect\citeauthoryear{Wang, Xu, and Sun}{Wang et~al\mbox{.}}{2022}]%
        {WangXS22}
\bibfield{author}{\bibinfo{person}{Yiyu Wang}, \bibinfo{person}{Jungang Xu}, {and} \bibinfo{person}{Yingfei Sun}.} \bibinfo{year}{2022}\natexlab{}.
\newblock \showarticletitle{End-to-End Transformer Based Model for Image Captioning}. In \bibinfo{booktitle}{\emph{Thirty-Sixth {AAAI} Conference on Artificial Intelligence, {AAAI} 2022, Thirty-Fourth Conference on Innovative Applications of Artificial Intelligence, {IAAI} 2022, The Twelveth Symposium on Educational Advances in Artificial Intelligence, {EAAI} 2022}}. \bibinfo{publisher}{{AAAI} Press}, \bibinfo{pages}{2585--2594}.
\newblock


\bibitem[\protect\citeauthoryear{Wen, Yang, Zhou, and Sun}{Wen et~al\mbox{.}}{2022}]%
        {wen_kdd_22}
\bibfield{author}{\bibinfo{person}{Qingsong Wen}, \bibinfo{person}{Linxiao Yang}, \bibinfo{person}{Tian Zhou}, {and} \bibinfo{person}{Liang Sun}.} \bibinfo{year}{2022}\natexlab{}.
\newblock \showarticletitle{Robust Time Series Analysis and Applications: An Industrial Perspective}. In \bibinfo{booktitle}{\emph{{KDD} '22: The 28th {ACM} {SIGKDD} Conference on Knowledge Discovery and Data Mining, 2022}}. \bibinfo{publisher}{{ACM}}, \bibinfo{pages}{4836--4837}.
\newblock


\bibitem[\protect\citeauthoryear{Woo, Liu, Kumar, Xiong, Savarese, and Sahoo}{Woo et~al\mbox{.}}{2024}]%
        {abs-2402-02592}
\bibfield{author}{\bibinfo{person}{Gerald Woo}, \bibinfo{person}{Chenghao Liu}, \bibinfo{person}{Akshat Kumar}, \bibinfo{person}{Caiming Xiong}, \bibinfo{person}{Silvio Savarese}, {and} \bibinfo{person}{Doyen Sahoo}.} \bibinfo{year}{2024}\natexlab{}.
\newblock \showarticletitle{Unified Training of Universal Time Series Forecasting Transformers}.
\newblock \bibinfo{journal}{\emph{CoRR}}  \bibinfo{volume}{abs/2402.02592} (\bibinfo{year}{2024}).
\newblock


\bibitem[\protect\citeauthoryear{Xu, Wu, Li, and Bouvry}{Xu et~al\mbox{.}}{2024}]%
        {abs-2401-00230}
\bibfield{author}{\bibinfo{person}{Jingjing Xu}, \bibinfo{person}{Caesar Wu}, \bibinfo{person}{Yuan{-}Fang Li}, {and} \bibinfo{person}{Pascal Bouvry}.} \bibinfo{year}{2024}\natexlab{}.
\newblock \showarticletitle{Transformer Multivariate Forecasting: Less is More?}
\newblock \bibinfo{journal}{\emph{CoRR}}  \bibinfo{volume}{abs/2401.00230} (\bibinfo{year}{2024}).
\newblock


\bibitem[\protect\citeauthoryear{Yan and Ouyang}{Yan and Ouyang}{2018}]%
        {YanO18}
\bibfield{author}{\bibinfo{person}{Hongju Yan} {and} \bibinfo{person}{Hongbing Ouyang}.} \bibinfo{year}{2018}\natexlab{}.
\newblock \showarticletitle{Financial Time Series Prediction Based on Deep Learning}.
\newblock \bibinfo{journal}{\emph{Wirel. Pers. Commun.}} \bibinfo{volume}{102}, \bibinfo{number}{2} (\bibinfo{year}{2018}), \bibinfo{pages}{683--700}.
\newblock


\bibitem[\protect\citeauthoryear{Yuan, Chen, Chen, Codella, Dai, Gao, Hu, Huang, Li, Li, Liu, Liu, Liu, Lu, Shi, Wang, Wang, Xiao, Xiao, Yang, Zeng, Zhou, and Zhang}{Yuan et~al\mbox{.}}{2021}]%
        {florence}
\bibfield{author}{\bibinfo{person}{Lu Yuan}, \bibinfo{person}{Dongdong Chen}, \bibinfo{person}{Yi{-}Ling Chen}, \bibinfo{person}{Noel Codella}, \bibinfo{person}{Xiyang Dai}, \bibinfo{person}{Jianfeng Gao}, \bibinfo{person}{Houdong Hu}, \bibinfo{person}{Xuedong Huang}, \bibinfo{person}{Boxin Li}, \bibinfo{person}{Chunyuan Li}, \bibinfo{person}{Ce Liu}, \bibinfo{person}{Mengchen Liu}, \bibinfo{person}{Zicheng Liu}, \bibinfo{person}{Yumao Lu}, \bibinfo{person}{Yu Shi}, \bibinfo{person}{Lijuan Wang}, \bibinfo{person}{Jianfeng Wang}, \bibinfo{person}{Bin Xiao}, \bibinfo{person}{Zhen Xiao}, \bibinfo{person}{Jianwei Yang}, \bibinfo{person}{Michael Zeng}, \bibinfo{person}{Luowei Zhou}, {and} \bibinfo{person}{Pengchuan Zhang}.} \bibinfo{year}{2021}\natexlab{}.
\newblock \showarticletitle{Florence: {A} New Foundation Model for Computer Vision}.
\newblock \bibinfo{journal}{\emph{CoRR}} (\bibinfo{year}{2021}).
\newblock


\bibitem[\protect\citeauthoryear{Zhang, Kishore, Wu, Weinberger, and Artzi}{Zhang et~al\mbox{.}}{2020}]%
        {bertscore}
\bibfield{author}{\bibinfo{person}{Tianyi Zhang}, \bibinfo{person}{Varsha Kishore}, \bibinfo{person}{Felix Wu}, \bibinfo{person}{Kilian~Q. Weinberger}, {and} \bibinfo{person}{Yoav Artzi}.} \bibinfo{year}{2020}\natexlab{}.
\newblock \showarticletitle{BERTScore: Evaluating Text Generation with {BERT}}. In \bibinfo{booktitle}{\emph{8th International Conference on Learning Representations, {ICLR} 2020}}. \bibinfo{publisher}{OpenReview.net}.
\newblock


\bibitem[\protect\citeauthoryear{Zhang, Deng, Liu, Pan, and Bing}{Zhang et~al\mbox{.}}{2023}]%
        {zhang2023}
\bibfield{author}{\bibinfo{person}{Wenxuan Zhang}, \bibinfo{person}{Yue Deng}, \bibinfo{person}{Bing Liu}, \bibinfo{person}{Sinno~Jialin Pan}, {and} \bibinfo{person}{Lidong Bing}.} \bibinfo{year}{2023}\natexlab{}.
\newblock \showarticletitle{Sentiment Analysis in the Era of Large Language Models: {A} Reality Check}.
\newblock \bibinfo{journal}{\emph{CoRR}}  \bibinfo{volume}{abs/2305.15005} (\bibinfo{year}{2023}).
\newblock
\urldef\tempurl%
\url{https://doi.org/10.48550/arXiv.2305.15005}
\showURL{%
\tempurl}


\bibitem[\protect\citeauthoryear{Zhang, Wei, and Zhou}{Zhang et~al\mbox{.}}{2019}]%
        {ZhangWZ19}
\bibfield{author}{\bibinfo{person}{Xingxing Zhang}, \bibinfo{person}{Furu Wei}, {and} \bibinfo{person}{Ming Zhou}.} \bibinfo{year}{2019}\natexlab{}.
\newblock \showarticletitle{{HIBERT:} Document Level Pre-training of Hierarchical Bidirectional Transformers for Document Summarization}. In \bibinfo{booktitle}{\emph{Proceedings of the 57th Conference of the Association for Computational Linguistics, {ACL} 2019}}. \bibinfo{publisher}{Association for Computational Linguistics}, \bibinfo{pages}{5059--5069}.
\newblock


\bibitem[\protect\citeauthoryear{Zhou, Zhang, Peng, Zhang, Li, Xiong, and Zhang}{Zhou et~al\mbox{.}}{2021}]%
        {ZhouZPZLXZ21}
\bibfield{author}{\bibinfo{person}{Haoyi Zhou}, \bibinfo{person}{Shanghang Zhang}, \bibinfo{person}{Jieqi Peng}, \bibinfo{person}{Shuai Zhang}, \bibinfo{person}{Jianxin Li}, \bibinfo{person}{Hui Xiong}, {and} \bibinfo{person}{Wancai Zhang}.} \bibinfo{year}{2021}\natexlab{}.
\newblock \showarticletitle{Informer: Beyond Efficient Transformer for Long Sequence Time-Series Forecasting}. In \bibinfo{booktitle}{\emph{Thirty-Fifth {AAAI} Conference on Artificial Intelligence, {AAAI} 2021, Thirty-Third Conference on Innovative Applications of Artificial Intelligence, {IAAI} 2021, The Eleventh Symposium on Educational Advances in Artificial Intelligence, {EAAI} 2021}}. \bibinfo{publisher}{{AAAI} Press}, \bibinfo{pages}{11106--11115}.
\newblock


\bibitem[\protect\citeauthoryear{Zhou, Niu, Wang, Sun, and Jin}{Zhou et~al\mbox{.}}{2023}]%
        {zhou2023onefitsall}
\bibfield{author}{\bibinfo{person}{Tian Zhou}, \bibinfo{person}{Peisong Niu}, \bibinfo{person}{Xue Wang}, \bibinfo{person}{Liang Sun}, {and} \bibinfo{person}{Rong Jin}.} \bibinfo{year}{2023}\natexlab{}.
\newblock \showarticletitle{One Fits All: Power General Time Series Analysis by Pretrained {LM}}. In \bibinfo{booktitle}{\emph{Advances in Neural Information Processing Systems 36: Annual Conference on Neural Information Processing Systems 2023}}.
\newblock


\end{thebibliography}
\appendix

\section{Using Different Percentages of Generated Data}
Table \ref{percentages} shows the performance of TSLM using different percentages of denoised generated data. We show that by just using 25\% of the total generated data, we are able to significantly improve the results compared to training the model with only the original data (0\% of denoised generated data). Then, the testing results keep improving when we use more generated data for training. This means that having a large quantity of accurate training pairs is the key to improve the generalization of our model. BERTScore for SYNTH reaches the highest value when using 50\% of the generated data. This phenomenon is understandable, because when we generate a very large quantity of data, the model will generalize better, but also will generate captions that are both accurate and different from the groundtruth captions in the testing set, and by consequence the evaluation metrics can start to decrease.  

\begin{table}[ht]
\small
\caption{Testing results using different percentages of denoised generated data.
\label{percentages}}
\centering
\begin{tabular}{||c|c|c|c|c||}
\hline
\textbf{Dataset} & \textbf{\% denoised data } & \textbf{R-L} & \textbf{BERTScore} & \textbf{TSLMScore} \\ \hline
\multirow{5}{*}{STOCK} &0  & 54.29  & 0.73&3.66\\
& 25 & 59.06 &0.75 &4.32\\&
 50  & 61.98  & 0.77&4.32\\&
 75&  64.24 & 0.79&\textbf{4.44}\\&
 100 & \textbf{66.45} & \textbf{0.80} & 4.42\\
 \hline
 \multirow{5}{*}{SYNTH} &0  & 75.12  & 0.82&6.20\\
& 25 & 81.49 &0.86 &6.48\\&
 50  &81.69 & \textbf{0.89} &6.93\\&
 75& 83.11  & \textbf{0.89}&\textbf{7.02}\\&
 100 & \textbf{83.20} & 0.88 & 6.98\\
 \hline
\end{tabular}
\end{table}

\begin{figure*}[ht!]
\centering
\includegraphics[width=1.0\textwidth]{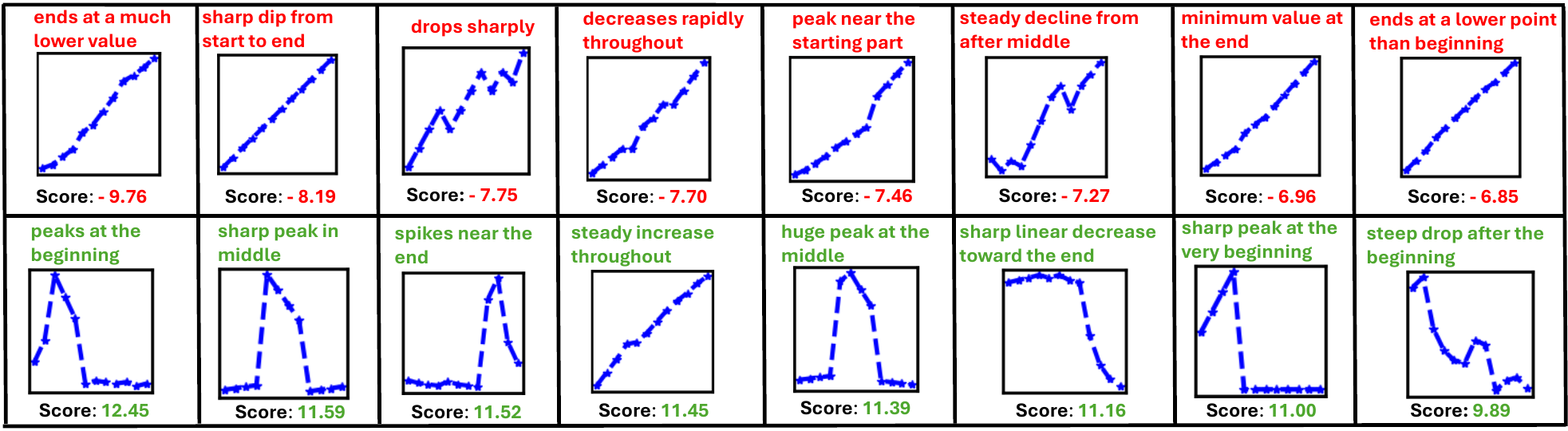}
\caption{Examples of generated time series-caption pairs with their predicted denoising scores computed from the cross-modal dense retrieval model. The first row represents noisy generated samples that are assigned a low score from the denoising model, and by consequence these samples are removed to denoise the generated data. The second row represents high-quality generated data that are assigned a high score from the denoising model, and by consequence these samples are kept.}
\label{denoising_examples}
\end{figure*}

\begin{figure*}[ht!]
\centering
\includegraphics[width=0.5\textwidth]{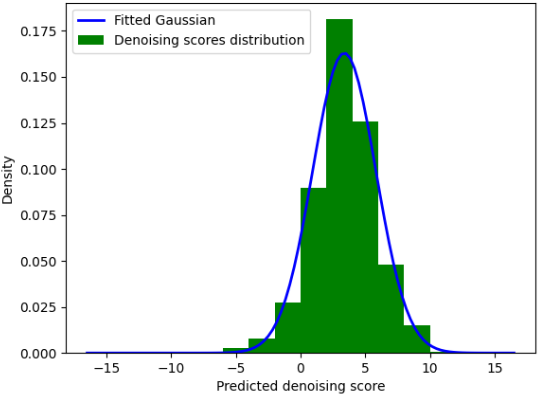}
\caption{Denoising scores distribution of the generated data. The distribution of denoising scores is approximated with a normal distribution with mean $\mu$ = 3.37 and standard deviation $\sigma$ = 2.44.}
\label{scores_dist}
\end{figure*}

\section{Examples of Denoising Scores of Generated Data}
\label{denoising_scores}
In Figure \ref{denoising_examples}, we show examples of generated time series-caption pairs with their predicted denoising score computed from the cross-modal dense retrieval model. In the first row, we show noisy generated samples that are assigned a low score from the denoising model. It is clear that the caption does not match the pattern of the time series, therefore these samples should be removed in order to denoise the generated data and avoid learning corrupted patterns. In the second row, we show clean generated examples that are assigned a large score from the denoising model. There is a strong agreement between the generated caption and the generated time series, and this strong agreement is translated into a large denoising score.  Our denoising model accurately detects the noisy generated samples which means that there is no need for human intervention to clean the generated data. As mentioned previously, the original size of the generated data is 203,554 and the denoising step results in removing 15,473 generated pairs leading to a total size of the denoised data that is equal to 188,081. TSLM learns from the original data and the denoised generated data to capture various patterns of time series captioning.

\begin{table*}[ht]
\small
\caption{Testing results using different denoising threshold $Th$ of the generated data.
\label{thresholds}}
\centering
\begin{tabular}{||c|c|c|c|c|c|c||}
\hline
\textbf{Dataset} & \textbf{Denoising threshold $Th$ } & \textbf{R-1} & \textbf{R-2} & \textbf{R-L} & \textbf{BERTScore} & \textbf{TSLMScore} \\ \hline
\multirow{7}{*}{STOCK} & $Th<min_{scores} = -9.76$ &62.99 &44.34 & 62.27  &0.76 &4.20\\
& $Th = -5$ &63.25 &44.79 &62.53  &0.77 &4.29\\
& $Th = -1$  &64.33 &46.87 &64.87   &0.77 &4.31\\
& $Th = 0$&\textbf{66.74} &\textbf{49.44} &\textbf{66.45}  &\textbf{0.80} &4.42\\
& $Th = 1$ &65.99 &48.24 &65.23 &\textbf{0.80}  &\textbf{4.46} \\
& $Th = 5$ &61.39 &43.86 &60.45 &0.77 &4.36 \\
& $Th>max_{scores} = 12.45$ &54.85 &37.89 &54.29 & 0.73&3.66\\
 \hline
 \multirow{7}{*}{SYNTH} & $Th<min_{scores} = -9.76$ &80.52 &65.95 & 77.57  &0.85 &6.63\\
& $Th = -5$ &83.44 &67.78 &81.68  &0.87 &6.71\\
& $Th = -1$  &83.68 &69.48 &81.93   &0.88 &6.83\\
& $Th = 0$&85.46 &71.43 &83.20  &\textbf{0.88} &\textbf{6.98}\\
& $Th = 1$ &\textbf{85.86} &\textbf{72.78} &\textbf{83.48} &0.87  &6.94 \\
& $Th = 5$ &80.99 &66.69 &79.24 &0.87 &6.84 \\
& $Th>max_{scores} = 12.45$ &75.43 &61.53 &75.12 &0.82 &6.20\\
 \hline
\end{tabular}
\end{table*}

\begin{figure*}
\centering
\begin{tabular}{cccc}
\includegraphics[width=0.26\textwidth]{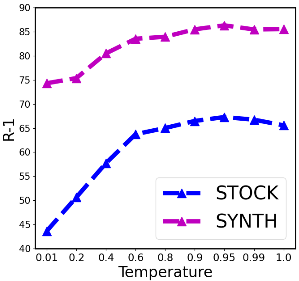} &
\includegraphics[width=0.26\textwidth]{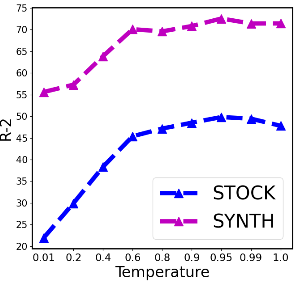} &
\includegraphics[width=0.26\textwidth]{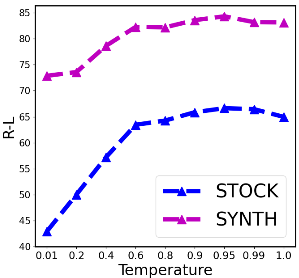} \\
(a) R-1 & (b) R-2 & (c) R-L  \\[6pt]
\end{tabular}
\begin{tabular}{cccc}
\includegraphics[width=0.26\textwidth]{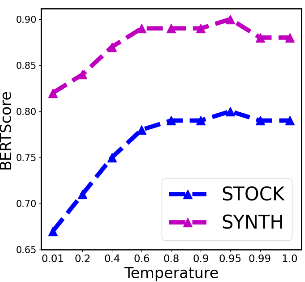} &
\includegraphics[width=0.26\textwidth]{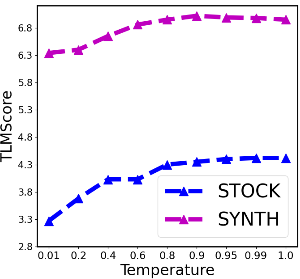} \\
(d) BERTScore  & (e) TSLMScore \\[6pt]
\end{tabular}
\vspace*{-3.25mm}
\caption{ Impact of the temperature on the evaluation metrics. In general, a higher temperature is preferable to generate multiple captions for the time series, and the best evaluation metrics are achieved for a temperature that is equal to 0.95.}\label{temperature}
\end{figure*}

\section{Analysis of Denoising Threshold $Th$}

In Figure \ref{scores_dist}, we show the distribution of the denoising scores of the generated data. The size of the generated data is 203,554 samples. The maximum score is 12.45 and the minimum score is -9.76. The distribution of scores is approximated with a normal distribution with mean $\mu$ = 3.37 and standard deviation $\sigma$ = 2.44.

We compare multiple denoising thresholds $Th$ by reporting the evaluation metrics on the testing sets after training TSLM with different denoised generated datasets based on the threshold $Th$. The results are highlighted in Table \ref{thresholds}. $Th<min_{scores} = -9.76$ means that all the generated data is used in the training of TSLM. $Th>max_{scores} = 12.45$ means that TSLM is trained without the generated data. $Th = -1$ is approximated to $\mu - 2 \times\sigma$ and $Th = 1$ is approximated to $\mu - \sigma$. The optimal results occur when the threshold $Th$ falls within the range of $[\mu - 2 \times\sigma$ , $\mu - \sigma]$, with a tendency towards $\mu - \sigma$. This indicates that trimming the left tail of the Gaussian distribution improves the quality of the denoised data, and by consequence leads to a more robust training of TSLM.

\section{Temperature Analysis}

\begin{figure*}[ht!]
\centering
\includegraphics[width=1\textwidth]{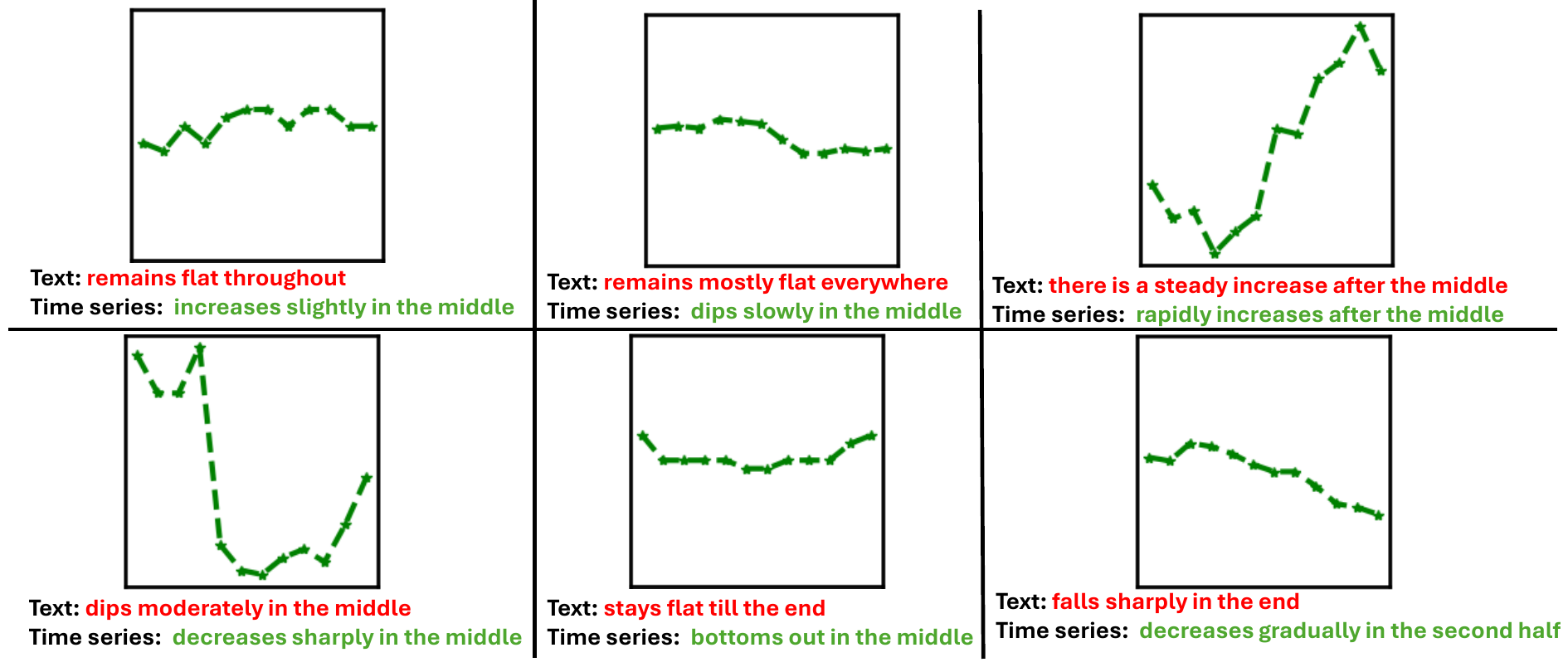}
\vspace*{-6mm}
\caption{Examples of generated captions from the STOCK dataset using TSLM (Text) and TSLM (TimeSeries). TSLM (TimeSeries) which is trained using the time series modality is able to generate more accurate captions about subtle variations in a time series such as slight vs rapid increase or decrease.}
\label{textvsembeddings}
\end{figure*}

The temperature in LLMs is an important hyperparameter that highly influences the model's output generation in terms of both the diversity and quality of the generated text. By adjusting the temperature, we enable a balance between the exploration of various possible outputs and the exploitation of the most likely outputs. With higher temperature values, LLMs are encouraged to generate multiple responses which is a useful aspect in applications that require creative and varied generated outputs. On the other hand, lower temperature values lead to more specific and high-probability tokens which is a useful aspect in tasks that need consistent generation such as translation and summarization. In the context of time series captioning, Figure \ref{temperature} shows the impact of the temperature on the evaluation metrics. In general, higher temperature is preferable to generate varied captions that describe multiple aspects and patterns in the time series. In the case of time series captioning, the input time series is ambiguous by nature and the high temperature values can enable TSLM to explore multiple interpretations and generate a variety of plausible captions. Therefore, diversity and novelty are desired aspects for time series captioning, and the high temperature values enhance the ability of TSLM for generating accurate and varied captions. As shown in Figure \ref{temperature}, the best evaluation metrics for both STOCK and SYNTH datasets are achieved for a temperature value that is equal to 0.95. Then, for extremely large temperature values (0.99, 1.0), we notice that the evaluation metrics start to decrease as these very high temperature values lead to selecting very unlikely tokens (overly exploring) which causes less accuracy and consistency in the generated captions.

\section{Text Embeddings vs Time Series Embeddings}

Figure \ref{textvsembeddings} shows examples of generated captions from the STOCK dataset using TSLM (Text) and TSLM (TimeSeries). The objective of this comparison is to show the importance of the time series modality compared to the text modality. TSLM (TimeSeries) which is trained using the time series modality is able to generate more accurate captions about subtle variations in a time series compared to TSLM (Text) which is trained with the text modality. For example, the first caption of the first row is described as \textit{remains flat throughout} by TSLM (Text). However, there is a very slight increase in the middle which is captured by TSLM (TimeSeries) as it generates the caption \textit{increases slightly in the middle}. For the third example in the first row, TSLM (Text) predicts that \textit{there is a steady increase after the middle}. However, the time series exhibits a rapid increase which is captured by  TSLM (TimeSeries) as it predicts the caption \textit{rapidly increases after the middle}. For the third example in the second row, TSLM (Text) prediction is \textit{falls sharply in the end}. However, the time series exhibits a gradual decrease and TSLM (TimeSeries) confirms this pattern by predicting \textit{decreases gradually in the second half}. These examples in addition to others reveal the importance of the time series embeddings in capturing subtle variations that are not possibly learned from the text modality.

\section{Limitations}
\label{limitations}
In this section, we highlight three limitations of TSLM:

\textbf{Larger datasets from various domains}: We showed the effectiveness of TSLM for handling the data scarcity problem. Training a domain-agnostic TSLM with larger datasets from various domains is the next step to study the effectiveness of TSLM in learning from multiple data sources.

\textbf{Qualitative evaluation of denoising generated data}: The denoising step is only evaluated qualitatively with manual inspection of the predicted scores of many generated time series-caption pairs as shown in Appendix \ref{denoising_scores}. It is infeasible to manually check the scores for all generated data to ensure every noisy sample is removed. However, our ablation study comparing our full model TSLM to TSLM (w/o denoising) shows the effectiveness of our denoising step in terms of reducing the noise level in the synthetically generated data.

\textbf{Static textual representation tagging}: We used the three phases tagging of a time series in this work as it is adequate with the ground truth captions. In general, different tagging can be used to segment the time series sequence, and TSLM can learn from various tagging strategies to generate captions that describe more phases than start, middle, and end. In addition, the caption can describe the exact location of patterns using the coordinates of the $x$ axis, and this necessitates explicitly adding the $x$ axis values into the text representation in addition to the $y$ values using the form of $x:y$. TSLM can be easily adapted to these cases when more diverse datasets are publicly available to reflect a wide range of scenarios in terms of ground truth captions.

\end{document}